\documentclass[preprint,onecolumn,amsmath,nofootinbib]{revtex4-2}

\usepackage[utf8]{inputenc}
\usepackage{graphicx, floatrow}
\usepackage[caption=false]{subfig}
\usepackage{amssymb, amsfonts, amsmath, dsfont}
\usepackage{comment}
\usepackage{multirow}
\usepackage{makecell}
\usepackage[T1]{fontenc}
\usepackage{floatrow}
\usepackage[table]{xcolor}
\usepackage{letltxmacro}
\usepackage[section]{placeins}

\usepackage[]{algorithm2e}

\usepackage{hyperref}

\DeclareRobustCommand{\Sec}[1]{Section~\ref{sec:#1}}

\DeclareRobustCommand{\Fig}[1]{Figure~\ref{fig:#1}}

\DeclareRobustCommand{\Tab}[1]{Table~\ref{tab:#1}}
\DeclareRobustCommand{\Tabs}[2]{Tables~\ref{tab:#1} and~\ref{tab:#2}}

\LetLtxMacro{\originaleqref}{\eqref}
\renewcommand{\eqref}[1]{\originaleqref{eq:#1}}

\providecommand\subsecref[1]{\ref{subsec:#1}}
\providecommand\secref[1]{\ref{sec:#1}}

\global\long\def\v#1{\boldsymbol{#1}}%

\global\long\def\dif#1{\mathrm{d}#1}%

\global\long\def\subtmo#1{#1_{\text{3mo}}}%
\global\long\def\subsmo#1{#1_{\text{6mo}}}%

\def\be{\begin{equation}} 
\def\ee{\end{equation}} 
\def\longrightharpoonup{\relbar\joinrel\rightharpoonup}
\def\longleftharpoondown{\leftharpoondown\joinrel\relbar}

\def\longrightleftharpoons{
  \mathop{
    \vcenter{
      \hbox{
      \ooalign{
        \raise1pt\hbox{$\longrightharpoonup\joinrel$}\crcr
	  \lower1pt\hbox{$\longleftharpoondown\joinrel$}
	  }
      }
    }
  }
}

\newcommand \bea {\begin{eqnarray}} 
\newcommand \eea {\end{eqnarray}}



\definecolor{ublue}{HTML}{007aff}

\begin{document}

% fix for footnote bug
% see https://tex.stackexchange.com/a/373701
\count\footins = 1000

\title{Modeling Disease Progression in Mild Cognitive Impairment and Alzheimer’s Disease with Digital Twins}

\author{Daniele Bertolini}
\email{daniele@unlearn.ai}
\affiliation{Unlearn.AI, Inc., 75 Hawthorne, San Francisco, CA 94105}

\author{Anton D. Loukianov}
\affiliation{Unlearn.AI, Inc., 75 Hawthorne, San Francisco, CA 94105}

\author{Aaron M. Smith}
\affiliation{Unlearn.AI, Inc., 75 Hawthorne, San Francisco, CA 94105}

\author{David Li-Bland}
\affiliation{Unlearn.AI, Inc., 75 Hawthorne, San Francisco, CA 94105}

\author{Yannick Pouliot}
\affiliation{Unlearn.AI, Inc., 75 Hawthorne, San Francisco, CA 94105}

\author{Jonathan R. Walsh}
\affiliation{Unlearn.AI, Inc., 75 Hawthorne, San Francisco, CA 94105}

\author{Charles K. Fisher}
\affiliation{Unlearn.AI, Inc., 75 Hawthorne, San Francisco, CA 94105}

\author{for the Critical Path for Alzheimer's Disease}
\thanks{Data used in the preparation of this article were obtained from the Critical Path for Alzheimer's Disease (CPAD). As such, the investigators within CPAD contributed to the design and implementation of the CPAD database and/or provided data, but did not participate in the analysis of the data or the writing of this report.}

\author{the Alzheimer's Disease Neuroimaging Initiative}
\thanks{Data used in preparation of this article were obtained from the Alzheimer’s Disease Neuroimaging Initiative (ADNI) database (\href{url}{adni.loni.usc.edu}). As such, the investigators within the ADNI contributed to the design and implementation of ADNI and/or provided data but did not participate in analysis or writing of this report. A complete listing of ADNI investigators can be found in \href{http://adni.loni.usc.edu/wp-content/uploads/how_to_apply/ADNI_Acknowledgement_List.pdf}{this document}.}

\date{\today}

\begin{abstract} 
Alzheimer's Disease (AD) is a neurodegenerative disease that affects subjects in a broad range of severity and is assessed in clinical trials with multiple cognitive and functional instruments.  As clinical trials in AD increasingly focus on earlier stages of the disease, especially Mild Cognitive Impairment (MCI), the ability to model subject outcomes across the disease spectrum is extremely important.  We use unsupervised machine learning models called Conditional Restricted Boltzmann Machines (CRBMs) to create Digital Twins of AD subjects. Digital Twins are simulated clinical records that share baseline data with actual subjects and comprehensively model their outcomes under standard-of-care.  The CRBMs are trained on a large set of records from subjects in observational studies and the placebo arms of clinical trials across the AD spectrum. These data exhibit a challenging, but common, patchwork of measured and missing observations across subjects in the dataset, and we present a novel model architecture designed to learn effectively from it. We evaluate performance against a held-out test dataset and show how Digital Twins simultaneously capture the progression of a number of key endpoints in clinical trials across a broad spectrum of disease severity, including MCI and mild-to-moderate AD.
\end{abstract}

\maketitle
%\tableofcontents

\section{Introduction}%
\label{sec:introduction}
The development of computational models that comprehensively and accurately forecast a patient's prognosis under the current standard-of-care has the potential to revolutionize medical research. Recall that a typical clinical trial compares the safety and efficacy of an investigational therapy to an established therapy, often in combination with a dummy treatment (i.e., placebo). Therefore, generative models trained to represent disease progression with existing treatments can be used to generate synthetic clinical records, which we call `Digital Subjects', for use in clinical trial design or as external control groups~\cite{walsh2020generating, fisher2019machine}. Similarly, conditional generative models can be used to generate distributions of potential control outcomes for individual patients (\cite{walsh2020generating, fisher2019machine,yoon2018ganite,walshstats2}), which we call `Digital Twins', that can be incorporated into clinical trials to increase statistical power or reduce required sample sizes~\cite{schuler2020increasing, walshstats2}.

Alzheimer's Disease (AD) is one indication with a particularly dire need for the development of new technologies that could increase the probability of finding an effective treatment. Despite decades of research, there is still no approved disease-modifying treatment for AD, and it often takes years to enroll enough subjects to achieve sufficient statistical power for a new clinical trial~\cite{cummings2014alzheimer, cummings2018price}. Therefore, the ability to decrease required sample sizes by augmenting AD clinical trials with data from generative models would make it possible to search for new AD treatments more quickly.

During a typical AD clinical trial, each subject reports to a clinical site at regular intervals (e.g., every 3 or 6 months) and is assessed with a variety of metrics to measure cognitive function and overall health. Usually, cognitive function is assessed through composite questionnaires such as the Alzheimer's Disease Assessment Scale - Cognitive Subscale (ADAS-Cog) \cite{rosen1984new}, the Clinical Dementia Rating (CDR) \cite{hughes1982new}, and the Mini-Mental State Exam (MMSE) \cite{folstein1975mini}. In addition, vital signs, blood tests, and biomarkers derived from magnetic resonance imaging (MRI) or positron emission tomography (PET) may be measured. That is, data collected in clinical trials typically correspond to a type of panel data~\cite{wooldridge2010econometric} -- a multidimensional array that describes how every subject's health data changes over the course of the trial.

Previously, Fisher et al. \citep{fisher2019machine} demonstrated that a type of generative model called a Conditional Restricted Boltzmann Machine (CRBM) can be trained using an adversary to generate panel data representative of control groups of AD clinical trials. In addition, they showed that CRBMs can also be used as conditional generative models to forecast potential outcomes for individual subjects. Here, we aim to extend this work by incorporating additional training data that includes patient populations with earlier stage disease, additional cognitive measures such as CDR, as well as additional biomarkers. These improvements are crucial for addressing the needs of a changing AD clinical trials landscape that is increasingly focused on earlier stages of the disease including subjects with Mild Cognitive Impairment (MCI)~\cite{sperling2011toward}.

Most clinical trials focus on relatively homogeneous populations, and trials in AD are no exception. Few individual studies span the spectrum from MCI to moderate AD. In addition, an average clinical trial in AD only includes around 400 subjects~\cite{cummings2014alzheimer, cummings2018price}. Therefore, it's necessary to integrate data from multiple interventional and observational studies in order to collect a sufficiently broad dataset. Indeed, we use clinical records from nearly 7,000 subjects -- covering close to 35,000 subject-visits -- integrated from 21 different studies. Unfortunately, these different studies did not all collect the same measurements or have subjects visit clinical sites with the same frequency. As a result, integrating these different studies solves the sample size problem, but creates a missing data problem. For example, the components of ADAS-Cog were typically measured in 3-month intervals, whereas the components of CDR were typically measured in 6-month intervals. To overcome this problem, we introduce a new approach that combines two CRBMs, each trained to represent a specific timescale, into a single generative model for creating Digital Subjects and Digital Twins for AD.

Using analyses on a held-out test set, we demonstrate that a generative model composed of two CRBMs accurately simulates subjects with MCI and AD under the current standard-of-care. That is, statistical properties computed from Digital Twins agree with statistical properties computed from actual subjects in the control arms of previous AD clinical trials, which is a necessary condition for using Digital Twins to increase the statistical power of future AD trials. Moreover, we  demonstrate that the updated model represents a substantial improvement over previously published work, particularly for subjects with MCI.

The manuscript is organized as follows. \Sec{short-methods} provides a brief description of our dataset, modeling approach, and how we test the model and compare it to previous work. \Sec{results} presents various comparisons of Digital Twins generated by the model to actual subject records in order to assess the quality of the generative model. Finally, \Sec{discussion} discusses implications of these results and future directions for research.  Appendices~\ref{sec:methods},~\ref{subsec:endpoints-covariates}, and~\ref{sec:crbm-summary} give additional details on the methods used, and Appendices~\ref{sec:additional-results-endpoints} and~\ref{sec:additional-results-marginals} give additional results beyond those discussed in the main text.

\section{Methods}
\label{sec:short-methods}

\subsection{Data}
Previously, Fisher et al. \cite{fisher2019machine} used a dataset with 1,909 subjects covering 44 clinical variables to train and evaluate a generative model for AD clinical trajectories. Here, we improve on this previously published work by training a generative model using a much larger dataset composed of 6,919 subjects covering 64 clinical variables across a broader set of domains, obtained by integrating data from 21 different studies. 

Data were obtained from the C-Path Online Data Repository for Alzheimer\textquoteright s Disease (CODR-AD) \citep{romero2009coalition, neville2015development} and the Alzheimer\textquoteright s Disease Neuroimaging Initiative (ADNI) \citep{mueller2005ways}. The former correspond to data from the control arms of previously completed AD clinical trials, whereas the latter is a large observational study of MCI and AD subjects.  These two datasets differ in a number of ways.  Data from CODR-AD is primarily from mild to moderate AD studies, with a visit interval of at most 3 months.  While CODR-AD data is rich in ADAS-Cog and MMSE measurements as well as safety data such as laboratory tests, it lacks some measurements prevalent in many current AD trials, particularly CDR and certain biomarker data.  ADNI data has a visit interval of 6 months or more and contains data for subjects across the entire disease spectrum, with a substantial fraction of data for MCI subjects\footnote{ADNI also includes cognitively normal cohorts, which are not used in our datasets.}.  ADNI includes a more diverse set of measurements with both CDR and biomarker data but without, for example, laboratory tests.

We combined these two sources and built an integrated dataset with a more complete set of variables than either individual dataset can provide. This integrated dataset is comprised of panel data for 6,919 subjects with 34,224 subject-visits in which approximately 25\% of subjects have MCI and the remainder have AD.  Subject-matter experts were consulted to determine variables of clinical significance and, based on data availability, 64 variables were included.  Most of the variables were measured in regular 3- or 6-month intervals. A few variables measured only at the first (baseline) visit were treated as background variables, i.e., properties of the clinical state of subjects at the start of the study. The variables are listed in \Tab{variables-condensed}. Additional details on the variables and the construction of the dataset are provided in Appendix~\subsecref{methods_data}.

Before training, the dataset was split in the ratio $0.5:0.2:0.3$ into training, validation, and test datasets, stratified by study. The test dataset was held out until the end of model development and used for all analyses shown in this work unless otherwise noted.

%%%
\begin{table*}[tp!]
\caption{Variables used in modeling AD progression.  Background variables are denoted in italics.  The group(s), 3-month and/or 6-month, that each variable is used with in modeling, based on the visit interval of the data and the relationship to other variables, are given.}

{\renewcommand{\arraystretch}{0.7}%
\resizebox{0.9\textwidth}{!}{
\begin{tabular}{|c|c||c|c|}
\hline
\textbf{Variable Name}              & \textbf{Group} & \textbf{Variable Name} & \textbf{Group} \\ \hline
ADAS cancellation          & 3   & Alanine aminotransferase   & 3 \\ \hline
ADAS commands              & 3, 6 & Alkaline phosphatase       & 3 \\ \hline
ADAS comprehension         & 3, 6 & Aspartate aminotransferase & 3 \\ \hline
ADAS construction          & 3, 6 & Cholesterol                & 3 \\ \hline
ADAS delayed word recall   & 3, 6 & Creatinine kinase          & 3 \\ \hline
ADAS ideational            & 3, 6 & Creatinine                 & 3 \\ \hline
ADAS naming                & 3, 6 & Eosinophils                & 3 \\ \hline
ADAS orientation           & 3, 6 & Gamma-glutamyl transferase & 3 \\ \hline
ADAS remember instructions & 3, 6 & Glucose                    & 3 \\ \hline
ADAS spoken language       & 3, 6 & Hematocrit                 & 3 \\ \hline
ADAS word finding          & 3, 6 & Hemoglobin                 & 3 \\ \hline
ADAS word recall           & 3, 6 & Hemoglobin A1C             & 3 \\ \hline
ADAS word recognition      & 3, 6 & Indirect Bilirubin         & 3 \\ \hline
MMSE attention             & 3, 6 & Lymphocytes                & 3 \\ \hline
MMSE language              & 3, 6 & Monocytes                  & 3 \\ \hline
MMSE orientation           & 3, 6 & Platelet                   & 3 \\ \hline
MMSE recall                & 3, 6 & Potassium                  & 3 \\ \hline
MMSE registration          & 3, 6 & Sodium                     & 3 \\ \hline
CDR community              & 6   & Triglycerides              & 3 \\ \hline
CDR home \& hobbies        & 6   & Heart rate                 & 3 \\ \hline
CDR judgement              & 6   & Diastolic blood pressure   & 3 \\ \hline
CDR memory                 & 6   & Systolic blood pressure    & 3 \\ \hline
CDR orientation            & 6   & Weight                     & 3 \\ \hline
CDR personal care          & 6   & Serious adverse events     & 3 \\ \hline
\emph{Baseline Age}               & 3, 6 & \emph{AChEI/Memantine use}    & 3, 6   \\ \hline
\emph{Sex}                        & 3, 6 & \emph{History of hypertension}    & 3, 6   \\ \hline
\emph{Years of education}         & 3, 6 & \emph{History of type-2 diabetes} & 3, 6  \\ \hline
\emph{ApoE $\epsilon$4 allele count}   & 3, 6 & \emph{Amyloid status}   & 3, 6  \\ \hline
\emph{Region (Europe)}            & 3   & \emph{CSF phosphorylated tau 181} & 3, 6 \\ \hline
\emph{Region (Northern America)}  & 3   & \emph{CSF total tau}              & 3, 6  \\ \hline
\emph{Region (Other)}             & 3   & \emph{Vitamin B12}                & 3, 6  \\ \hline
\emph{Height}                     & 3, 6 & \emph{RCT or Observational Study} & 3, 6  \\ \hline

\end{tabular}
}}
\label{tab:variables-condensed}
\end{table*}%
%%%

\subsection{Modeling Approach}

Modeling the integrated panel dataset presents two challenges: variables are observed with two characteristic timescales (3-months and 6-months) and only a subset of those features are ever observed simultaneously.

There are at least three different ways to solve the multiple timescale problem. First, one could downsample all of the data to a 6-month visit interval. However, this makes the data unsuitable for many clinical trial applications, which typically record observations every 3 months. Alternatively, one could directly train a model with a 3-month visit frequency on the combined dataset. Training such a model is extremely challenging as some key variables, such as the components of CDR, would be missing in a large fraction of observations. Therefore, we adopt a third approach in which we explicitly construct a generative model with two timescales.

We handle the two-timescale character of the data explicitly by training a composite model consisting of two CRBMs of the form of \citet{walsh2020generating}: \emph{3-month} and \emph{6-month} CRBMs that model variables with those respective time intervals in the panel data.  Variables available at 3-month intervals are modeled by the 3-month CRBM, and variables \emph{only} available at 6-month intervals as well as key variables available at 3-month intervals are modeled by the 6-month CRBM.  Hence, there is a set of variables included in both CRBMs.  In our case, these shared variables are components of the ADAS-Cog and MMSE assessments as well as background variables.  The components of CDR are modeled by the 6-month CRBM alone.  \Tab{sample-counts} gives the number of subjects from each dataset used in the 3-month and 6-month CRBMs.  A summary of CRBM models is presented in Appendix~\ref{sec:crbm-summary}.

%%%
\begin{table*}[tp!]
\caption{The number subjects from each dataset used for each CRBM.  The 3-month CRBM uses all 6,919 subjects in the integrated dataset.}
\begin{tabular}{|c|c|c|}
\hline
\multirow{2}{*}{\textbf{Dataset}} & \multicolumn{2}{c|}{\textbf{Number of subjects}} \\ \cline{2-3} 
                                  &  \textbf{ 3-month CRBM } & \textbf{ 6-month CRBM } \\ \hline
CODR-AD  & 5331   & 208  \\ \hline
ADNI     & 1588   & 1588 \\ \hline
\end{tabular}
\label{tab:sample-counts}
\end{table*}%
%%%

The 3-month and 6-month CRBMs are naturally integrated into a single model.  Each can be separately trained, but when they are used to generate Digital Subjects or Digital Twins the 3-month CRBM is used first, followed by the 6-month CRBM.  Any shared variables are generated by the 3-month CRBM and conditioned upon (held fixed) when generating data from the 6-month CRBM.  This structure is natural, as the 3-month CRBM predicts a given subject's clinical trajectory for all variables except CDR.  Subsequently, key data variables generated by the 3-month CRBM are used by the 6-month CRBM to better inform the predictions of CDR on a 6-month timescale.

The 3-month and 6-month CRBMs are trained independently, each following a hyperparameter selection procedure presented in \citet{walsh2020generating} and reviewed here.  Using the training component of the dataset, a set of CRBMs are trained via stochastic gradient descent, each with different hyperparameters.  The validation dataset is used to compute metrics that judge model performance, and optimal hyperparameters are selected.  Finally, the CRBM is retrained using these optimal hyperparameters on both the training and validation components of the dataset.  This results in 3-month and 6-month CRBMs that can be integrated into the final model.

%%%
\begin{figure*}[tp!]
\includegraphics[width=0.9\columnwidth]{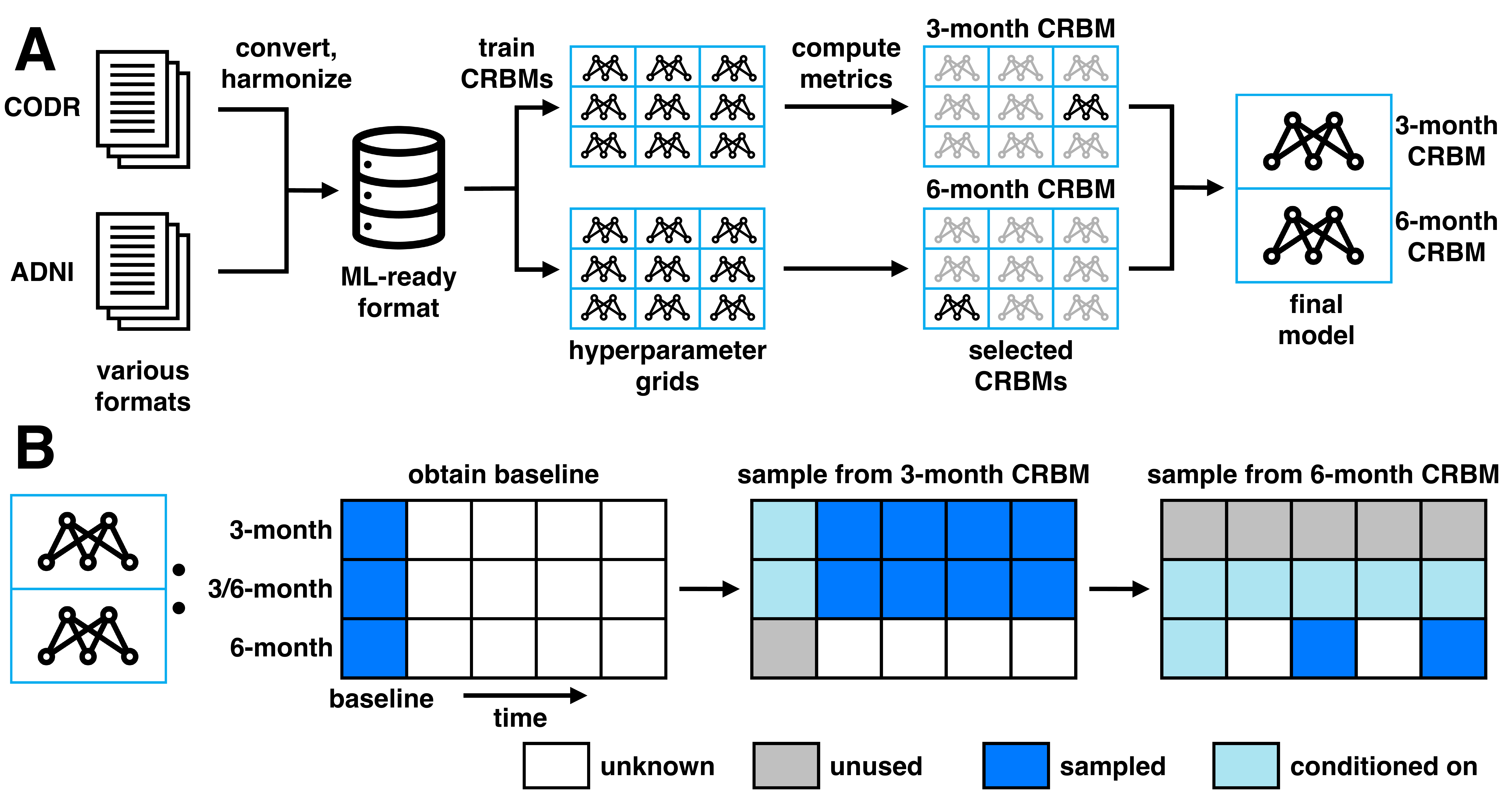}
\caption{{\bf An overview of the data curation, CRBM training, and model sampling}.  In (A), the data used to build the conditional generative model is curated from two sources, CODR-AD and ADNI.  The 3-month and 6-month CRBMs are built by training models in separate grids and using a selection process to determine optimal hyperparameters.  After retraining using these selected hyperparameters, the CRBMs are composed into the final model.  In (B), Digital Twins are built from baseline data by first sampling from the 3-month model then sampling from the 6-month model.  When sampling from the 6-month model, the variables shared with the 3-month model are conditioned upon using the samples generated from the 3-month model.
\label{fig:schematic}}
\end{figure*}
%%%

Data curation, CRBM training, and model sampling procedures are outlined in \Fig{schematic}.  For an extended description of the methods described here, see Appendix~\secref{methods}.

\subsection{Model Evaluation}

The goal of our modeling approach is to be able to generate Digital Twins whose panel data are \emph{well-calibrated}, meaning they accurately predict the probability distribution of actual subject data.  This is practically useful because it provides an unbiased prediction for subject outcomes along with an unbiased estimate for their variability.  Digital Twins go beyond simple predictions of mean outcomes -- for each subject, a set of Digital Twins predicts the distribution of possible outcomes.  Well-calibrated Digital Twins can be used to reliably predict subject-level or study-level outcomes and their expected variability.

We use the test dataset, which the constituent CRBMs have not seen, and compare test subjects to their Digital Twins created from the baseline data.  A number of different approaches are used to compare these two groups of data to assess variable-level, subject-level, and population-level measures.  As most outcomes of interest for clinical applications are linear combinations of the variables modeled, such as the ADAS-Cog total score, MMSE total score, or CDR Sum-of-Boxes (CDR-SB) score, we are particularly interested in the performance in modeling linear combinations of variables.

\section{Results}
\label{sec:results}

\subsection{Evaluating the Performance of the Model in Generating Digital Twins}
\label{subsec:results_twins}

We evaluate the ability of the trained conditional generative model to create well-calibrated forecasts for variables in three ways. First, for each variable, we ask if observed values tend to fall in the bulk of the distribution predicted by the model, conditioned on the observed data at time zero (the baseline data). Next, we ask if the population average means, variances, and pairwise (auto-)correlations predicted by the model are consistent with those estimated from the data. Finally, we train a linear classifier to distinguish between actual subject records and Digital Twins sampled from the model. All analyses are performed on the held-out test dataset not used by the model during training. Additional details on the methods used in this section can be found in Appendix~\secref{model_evaluation_methods}, and additional results on the statistical agreement between moments and correlations in disease progression measures for actual control subjects and their Digital Twins are presented in Appendices~\ref{sec:additional-results-endpoints} and~\ref{sec:additional-results-marginals}.

\begin{figure*}
\includegraphics[width=\columnwidth]{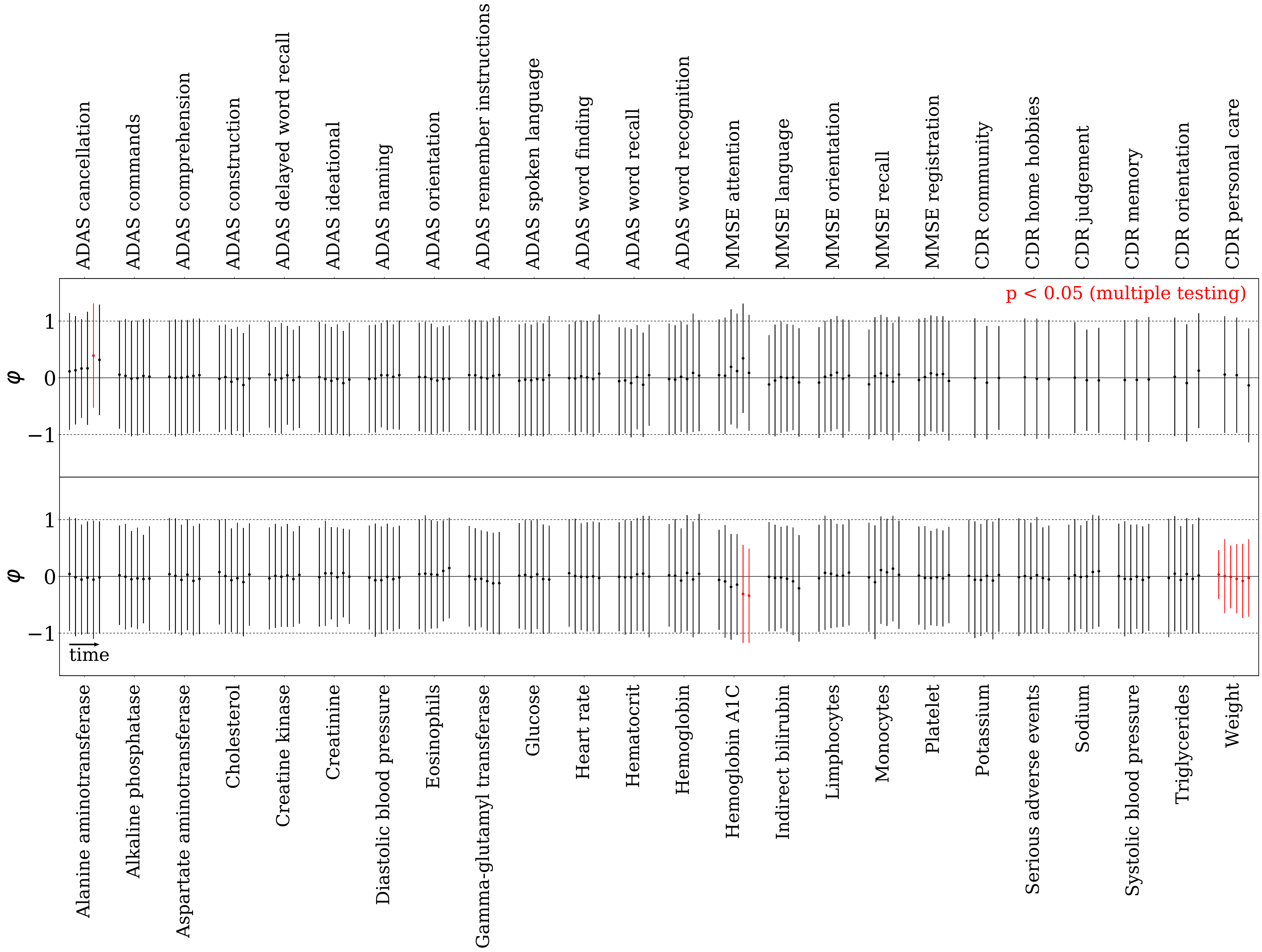}
\vspace{-0.1cm}
{\caption{{\bf Time-dependent marginal distributions computed from the model are well-calibrated.} For each variable, at each visit, we compute a statistic, $\varphi$, that has a standard normal distribution if observations from actual subjects are consistent with the distribution defined by the Digital Twins (see \subsecref{results_twins}). We plot the mean as a point and standard deviation as an error bar, and apply a Kolmogorov-Smirnov test to check if the distributions of $\varphi$ are consistent with $\mathcal{N}(0, 1)$.  Red marks any deviations that remain statistically significant after a Bonferroni correction. Means above (below) 0 indicate a biased model distribution with lower (higher) mean than the mean of the data; similarly, variances smaller (larger) than 1 indicate a model variance larger (smaller) than variance of the data.
\label{fig:zscores}}}
\end{figure*}

Our first analysis focuses on the calibration of the time-dependent marginal distributions, conditioned on observed baseline data for each subject. We generate 100 Digital Twins for each subject and, for each variable at each visit for each subject, verify the consistency of the distribution over these Digital Twins with the observed value from the actual subject.  In particular, for each variable at each visit, we compute the probability that a Digital Twin sampled from the conditional generative model has a larger value of that variable than the what was observed for a given subject. Then, we normalize this `p-value' by computing a transformation, $\varphi=\Phi^{-1}(p)$, in which $\Phi$ is the cumulative distribution function of the normal distribution. Finally, we perform a statistical test to determine if the distribution of $\varphi$ is consistent with a standard normal distribution, $\mathcal{N}(0, 1)$, which would be expected if the Digital Twins and actual subject data are drawn from the same distribution. \Fig{zscores} shows the mean and standard deviation of $\varphi$ for all variables and time points. For the vast majority of variables, differences between observed and expected distributions are not statistically significant after correcting for multiple testing. The only differences of note are that the variance for \emph{weight} predicted by the model is too large, and the mean values for \emph{hemoglobin A1C} at 15 and 18 months and \emph{ADAS cancellation} at 15 months slightly biased. Therefore, the time-dependent marginal distributions are generally well-calibrated.

\begin{figure*}
  \centering
  \raisebox{-0.5\height}{\includegraphics[height=3.3in]{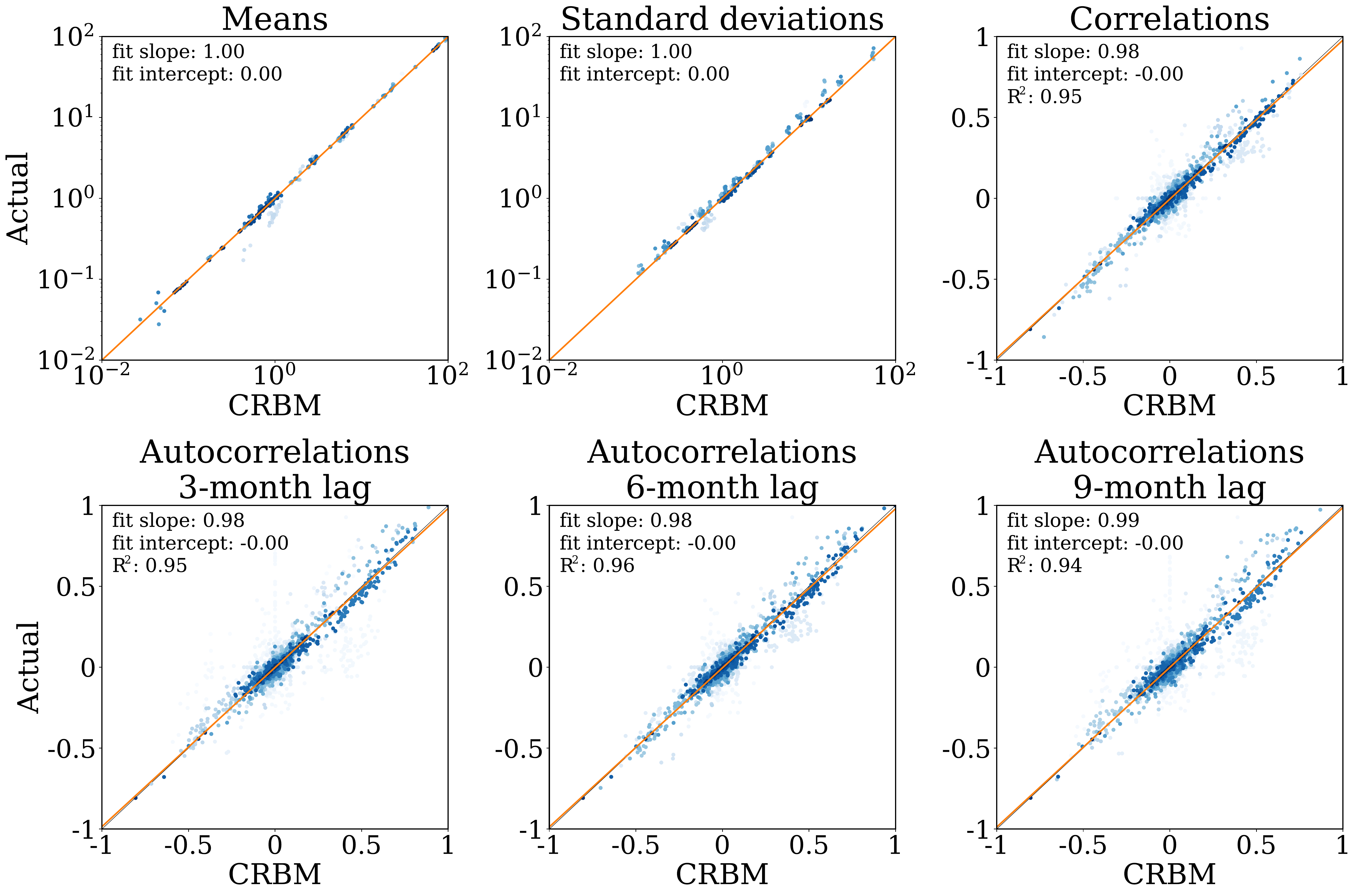}}
  \hspace*{.2in}
  \raisebox{-0.5\height}{\includegraphics[height=2in]{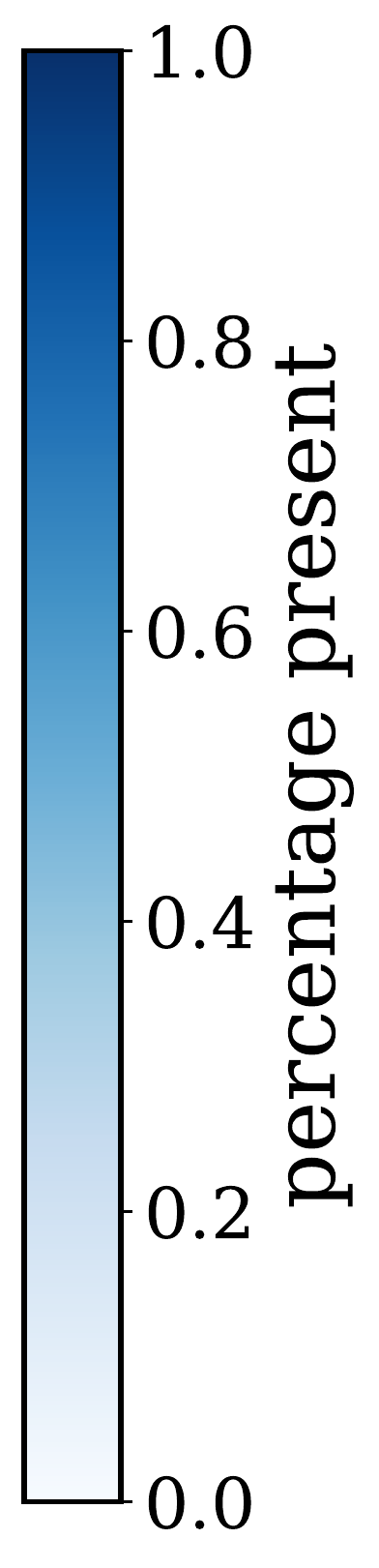}}
\caption{{\bf The model captures the leading statistical moments of the data.}  We compare the leading statistical moments of the distribution of all variables from the test data and from the model. A single Digital Twin is randomly generated for each subject and the moments are compared to the data for each 3-months follow-up visit up to 18 months. We report slope and intercept from a regression of the data against the model predictions, weighted by the fraction of data present at each particular point (indicated by the color bar on the right). For correlations and autocorrelations we also report the coefficient of determination $R^2$.
\label{fig:moments}}
\end{figure*}

Next, we examine on the ability of the conditional generative model to forecast the population-level statistics that are most important for linear combinations of variables such as the ADAS-Cog, CDR-SB, or MMSE composite scores. We compare population averaged means, variances, pairwise correlations, and the 3-, 6-, and 9-month lagged pairwise autocorrelations computed from Digital Twins and actual subjects. Specifically, we sampled a single Digital Twin for each subject in the test dataset to create a Digital Twin cohort. Then, we computed the above mentioned statistics from the Digital Twin cohort and from the test dataset. \Fig{moments} shows the values of these statistics from the test dataset plotted against the corresponding values computed from the Digital Twin cohort, with the goodness-of-fit assessed across statistics of the same type using linear regressions. In the regressions, points receive a weight proportional to the fraction of data present when computing that statistic to account for the impact of missing data.  Theil-Sen regression is used to quantify goodness-of-fit for the means and standard deviations to mitigate outlier dependence due to different units (and scales) across variables.  We find that all slopes are close to 1 and all intercepts are close to 0, illustrating that our model captures the leading statistical moments that are relevant for linear combinations of variables.  

%%%
%%%

\begin{figure*}
\includegraphics[width=0.8\columnwidth]{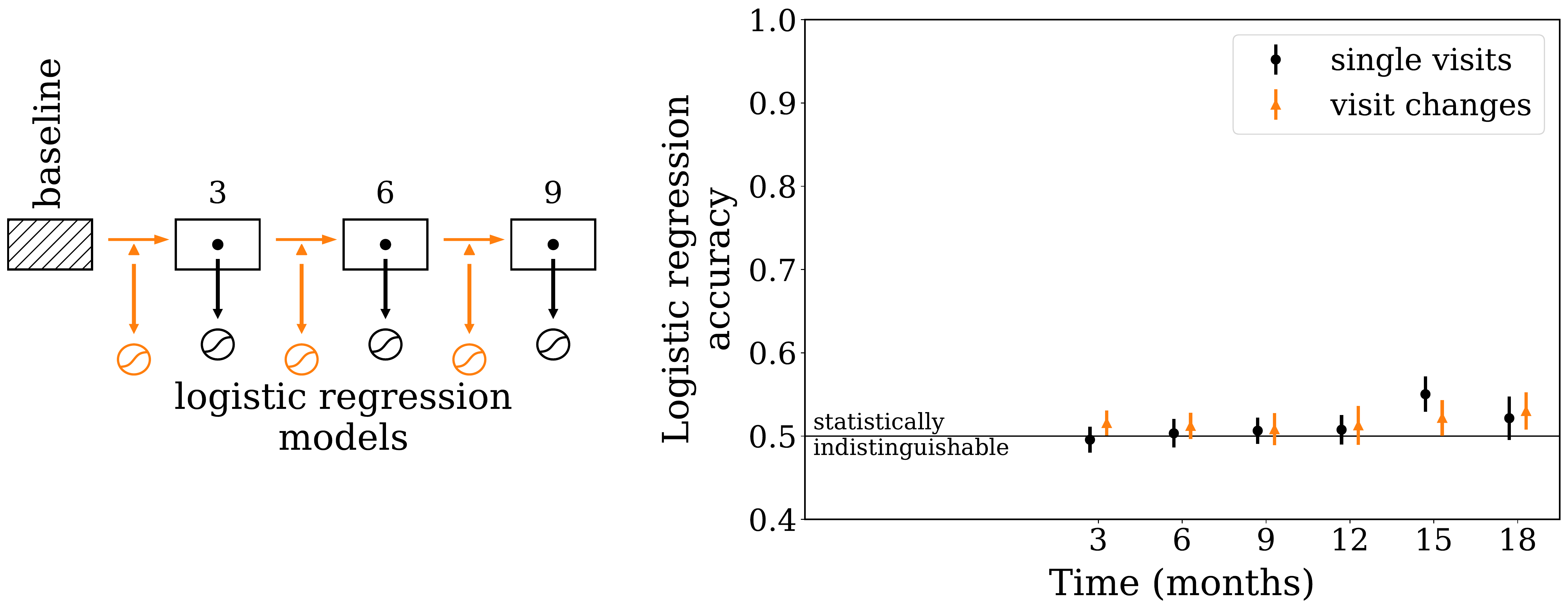}
\includegraphics[width=0.8\columnwidth]{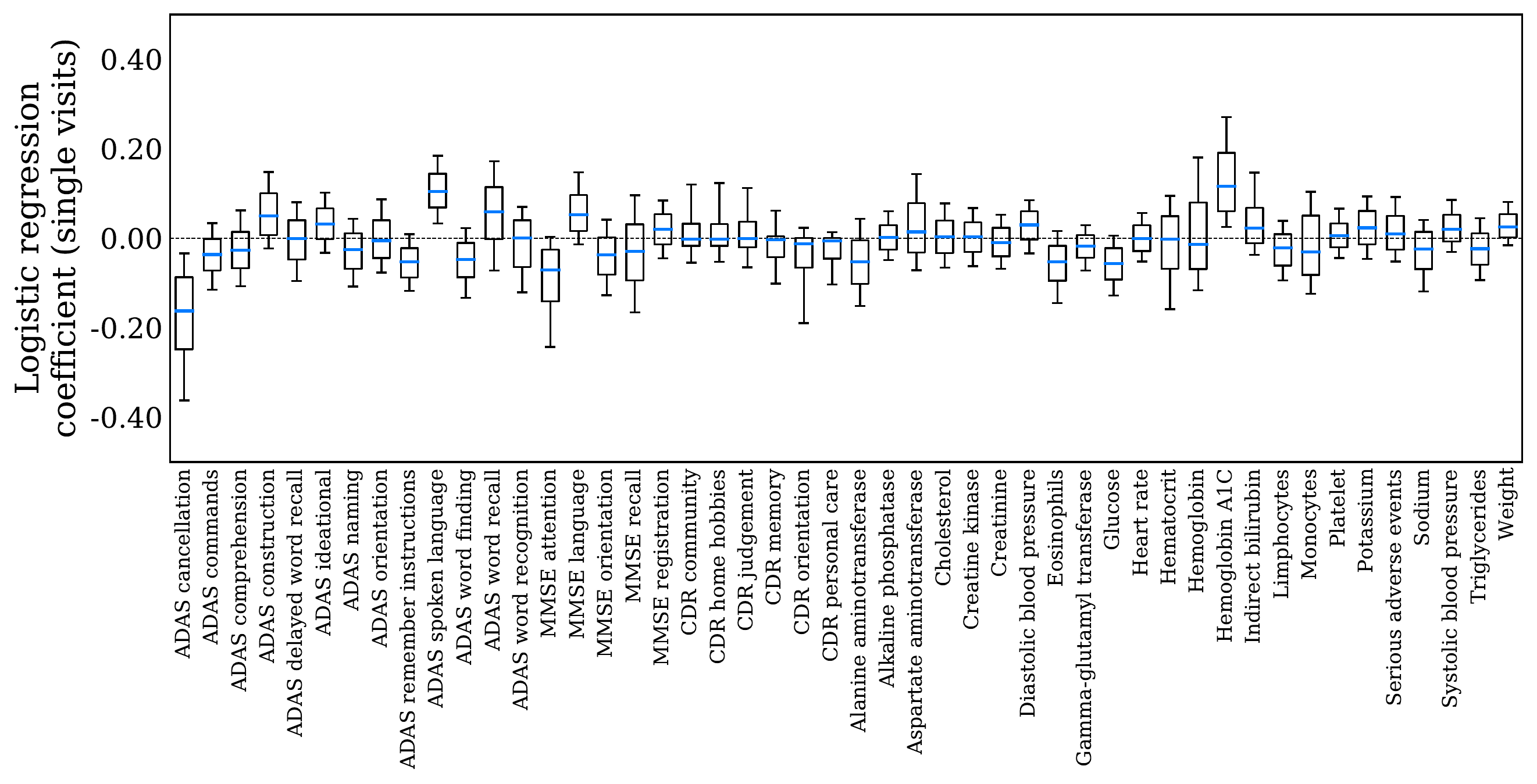}
\caption{{\bf Linear classifiers are unable to distinguish actual subjects from their Digital Twins.} We train a logistic regression model to differentiate actual subjects from their Digital Twins. Top panel: we compare two models, one in which we use data at each visit, and one in which we use differences between two consecutive visits. On the right, we report the accuracy of both models. An accuracy of 0.5 is consistent with random chance. Error bars indicate 95\% confidence interval errors. Bottom panel: boxplots for the distribution of logistic regression coefficients for the single visit model, where a weight equal to 0 corresponds to an irrelevant feature. 
\label{fig:adversary}}
\end{figure*}

Finally, we test the ability of a linear classifier to distinguish between actual subjects and their Digital Twins. In a sense, this is directly testing if Digital Twins are statistically indistinguishable from actual subjects, and is closely related to the adversarial methods used while training the model. For each time point, we train a logistic regression to distinguish between each subject and their Digital Twins. In addition, we train a logistic regression to distinguish between actual subjects and their Digital Twins using the difference in panel data between two consecutive time points. \Fig{adversary} shows that the linear classifiers only obtain accuracies that are consistent, or nearly consistent, with random chance. This demonstrates that individual subjects are statistically indistinguishable from their Digital Twins using linear combinations of variables.

%%%
%%%

\subsection{Progression of Common Clinical Endpoints}
\label{subsec:results_endpoints}

\begin{figure*}
\includegraphics[width=\columnwidth]{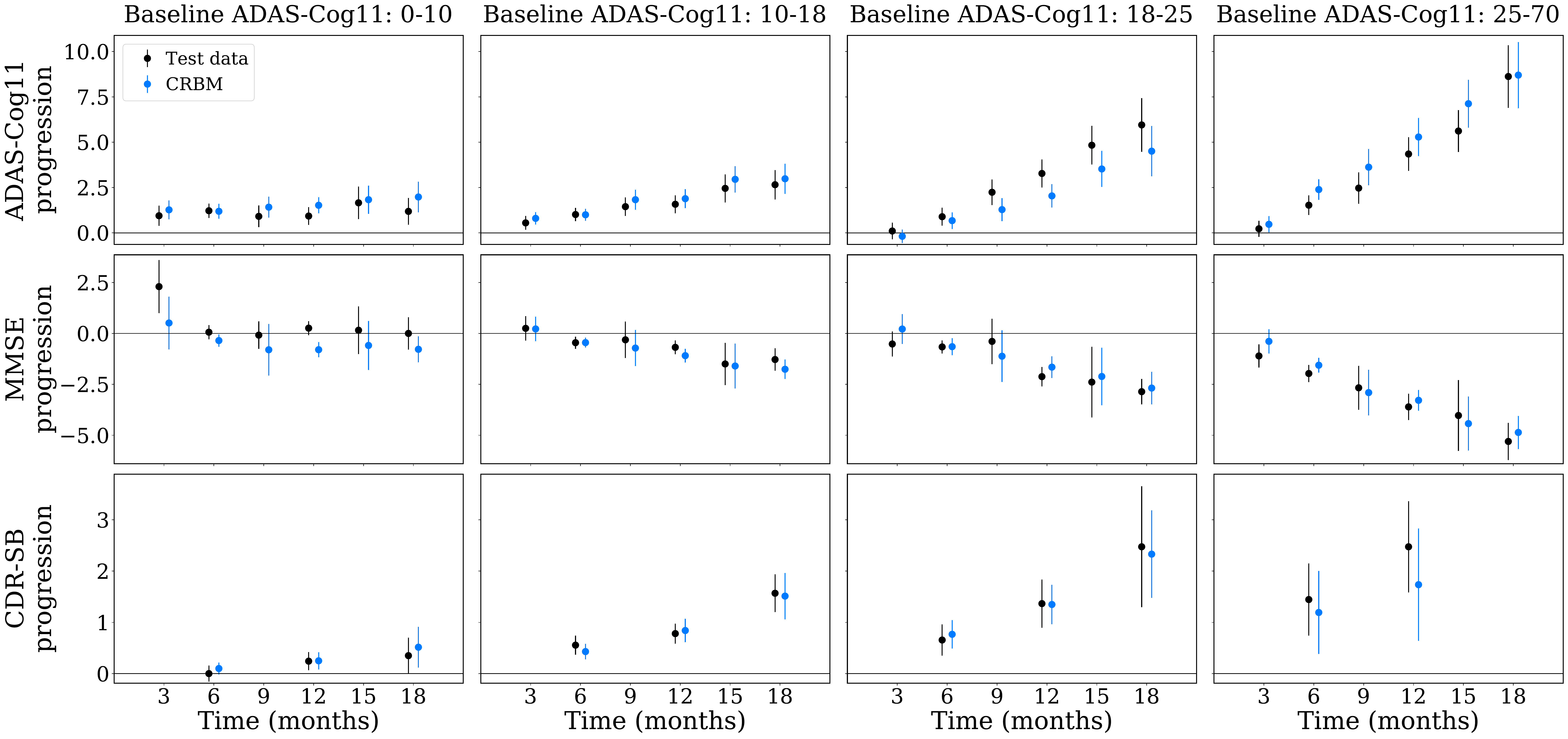}
\caption{{\bf Progression of clinical endpoints stratified by disease severity at baseline.} 
We compare the predicted mean change-from-baseline for ADAS-Cog11, CDR-SB, and MMSE to the mean change-from-baseline in test dataset. We stratified the population by ADAS-Cog11 measured at baseline, with lower values corresponding to milder cases. For each time point, the mean change-from-baseline, and its standard error, were estimated by drawing 100 Digital Twins for each subject and averaging over the population. We consistently predict the mean progression across all time points, scores, and cohorts. Error bars represent 95\% confidence interval errors. No CDR test data was available at 18 months for the most severe group.
\label{fig:endpoints}}
\end{figure*}

The components of the cognitive questionnaires are particularly important for clinical trial applications because ADAS-Cog, CDR, and MMSE are frequently used as inclusion criteria and endpoints in AD clinical trials. We model 13 components of the ADAS-Cog score, 6 components of the CDR score, and 5 components of the MMSE score, which are listed in \Tabs{covariates-child}{covariates-parent-only} (see Appendix~\subsecref{endpoints-covariates} for more detail). We compared the mean change-from-baseline in ADAS-Cog11, CDR-SB, and MMSE predicted from a Digital Twin cohort sampled from the conditional generative model to the corresponding changes estimated from the test dataset. We present the results stratified by disease severity at baseline. \Fig{endpoints} shows that we consistently predict the mean progression across all time points, scores, and cohorts, indicating that the conditional generative model is well-calibrated for these important clinical outcomes. Further, Appendix~\secref{additional-results-endpoints} shows that predicted changes-from-baseline for these scores in individual subjects are highly correlated with observed change from baseline from actual subjects. Taken together, these results demonstrate that the model can be used to accurately forecast the prognosis of subjects with MCI and AD.

%%%
%%%

\subsection{Comparison to an Earlier AD Progression Model}
\label{subsec:results_compare_old}
Previously, \citet{fisher2019machine} developed a machine learning model to simulate AD progression using a smaller dataset than that used in this manuscript. In addition, the prior model did not include the components of CDR, or other key biomarker and safety data, and mostly modeled subjects with mild to moderate AD. In this paper, we combined multiple data sources to create a larger and more comprehensive dataset. In addition, a much larger fraction of subjects in the new dataset had MCI, which resulted to a dramatic improvement in predictions for this cohort of subjects. We compare the predictions for ADAS-Cog11 progression from the two models on AD and MCI cohorts in \Fig{compare_ad_vs_mci}. In addition, \Tabs{old_model_marginals}{old_model_marginals_2} in Appendix~\secref{additional-results-marginals} compare the mean and standard deviation of the marginal distributions of each variable computed from the two models to those computed from the test dataset. The performance of the current model on subjects with AD is comparable to the previous model, but it substantially outperforms the previous model for subjects with MCI. 

%%%
\begin{figure*}[tp!]
\includegraphics[width=\columnwidth]{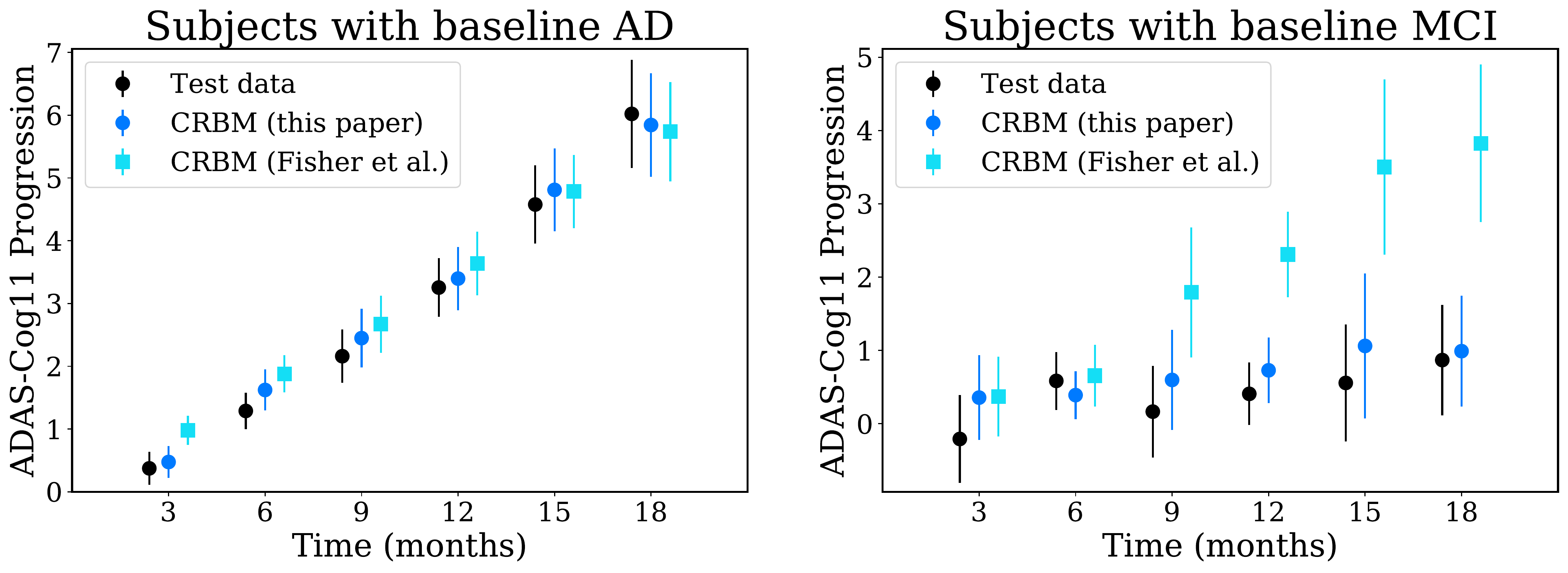}
\caption{{\bf Mean change-from-baseline in ADAS-Cog11 for subjects with AD or MCI.} 
We compare the predictions for mean ADAS-Cog11 progression from the CRBM model of \citet{fisher2019machine} and the model of this paper on two different cohorts: subjects with baseline diagnosis of AD or MCI. The model presented here outperforms the previous model on the MCI cohort. Mean progressions are calculated as in \Fig{endpoints} and error bars represent 95\% confidence interval errors.
\label{fig:compare_ad_vs_mci}}
\end{figure*}
%%%

\section{Discussion}
\label{sec:discussion}

A typical clinical trial compares the safetey and efficacy of an investigational therapy to a placebo and standard-of-care. Therefore, there are many uses for models that can generate accurate forecasts for disease progression under standard-of-care ranging from clinical trial design to statistical analysis plans that directly incorporate prognostic forecasts to improve statistical power \cite{schuler2020increasing,walshstats2}. Here, we built on previous work by training a generative model to forecast disease progression for subjects with AD, aiming to obtain well-calibrated forecasts across the spectrum of severity from early to severe disease.

A subject's prognosis can be forecast by using the model to generate Digital Twins -- longitudinal trajectories sampled from a multivariate probability distribution that describes how that subject's characteristics are likely to change after baseline. Each Digital Twin has 16 background variables that are only measured at baseline, and 48 time-dependent variables that are measured in 3- or 6-month intervals. In total, these 64 variables include cognitive questionnaires such as ADAS-Cog, CDR, and MMSE that are commonly used to assess disease progression in clinical trials, as well as lab tests and other biomarkers.

Modeling this comprehensive set of variables required integrating two large datasets of AD clinical trials and observational studies. However, the different studies in the integrated dataset typically observed patients in either 3- or 6-month intervals, and they did not always measure the same set of variables. Therefore, we introduced a new architecture comprised of two Conditional Restricted Boltzmann Machines (CRBMs), one to represent the 3-month timescale, and the other to represent the 6-month timescale. The two CRBMs can be combined to generate trajectories that have the variables measured in 3-month intervals and 6-month intervals. The challenge of modeling panel data with multiple timescales is commonly encountered when aggregating data from multiple studies, and the approach described could be easily extended to more general cases. Therefore, we expect this approach may be effective in modeling disease progression for many diseases.

We showed that, at the single variable level, actual subjects' observed measurements are consistent with the distribution of their Digital Twins at different visits. At the population level, we showed that Digital Twins capture the leading statistical moments of the data, including means, standard deviations, correlations, and lagged autocorrelations of all variables.  Finally, we have shown that logistic regression cannot distinguish actual subjects from their Digital Twins, with an accuracy consistent or nearly consistent with random chance. We also showed that the model accurately predicts the progression of ADAS-Cog11, CDR-SB, and MMSE, across multiple severity cohorts. Therefore, this generative model provides reliable forecasts for disease progression of the majority of patient characteristics that are of interest in AD clinical trials.

In comparison to a previous model used to forecast disease progression in subjects with AD presented by Fisher et al. \cite{fisher2019machine}, the model described here was based on a dataset with more than 3 times as many subjects, as well as additional clinical variables. In addition to providing more comprehensive forecasts, the new model provides more accurate forecasts -- particularly for patients with early stage disease. 

Generative models have a number of potential applications when applied to health data. For example, a generative model can be used to create datasets that have desired statistical properties but don't contain any private health information. This application could be useful for training discriminative models while protecting patient privacy or intellectual property rights of data owners. Similarly, cohorts of Digital Subjects can be used for simulating clinical trials with different inclusion criteria to facilitate clinical trial design.  

Similarly, conditional generative models have a number of potential applications in clinical trials, and even clinical care. For example, Digital Twins can be integrated into clinical trials as prognostic scores to increase their power or reduce the number of subjects required to achieve a desired power \cite{schuler2020increasing, walshstats2}. Further into the future, conditional generative models that can create comprehensive forecasts of likely outcomes for individual patients could form the basis of clinical decision support systems.

Here, we have demonstrated that a particular type of generative model (i.e., CRBMs) can be used to accurately model disease progression for patients with MCI or AD. CRBMs are particularly useful for this task because they effectively model panel data, can easily handle binary, continuous, or ordinal data types, can impute missing data at train or test time, and can be used as a generative model (to create Digital Subjects) or as a conditional generative model (to create Digital Twins). Future work could explore the advantages and disadvantages of different types of generative models for these types of data.

\section{Data Availability}

Certain data used in the preparation of this article were obtained from the Alzheimer’s Disease Neuroimaging Initiative (ADNI) database (\href{url}{adni.loni.usc.edu}). The ADNI was launched in 2003 as a public-private partnership, led by Principal Investigator Michael W. Weiner, MD. The primary goal of ADNI has been to test whether serial magnetic resonance imaging (MRI), positron emission tomography (PET), other biological markers, and clinical and neuropsychological assessment can be combined to measure the progression of mild cognitive impairment (MCI) and early Alzheimer’s disease (AD). For up-to-date information, see \href{url}{www.adni-info.org}.

Certain data used in the preparation of this article were obtained from the Critical Path for Alzheimer's Disease (CPAD) database. In 2008, Critical Path Institute, in collaboration with the Engelberg Center for Health Care Reform at the Brookings Institution, formed the Coalition Against Major Diseases (CAMD), which was then renamed to CPAD in 2018. The Coalition brings together patient groups, biopharmaceutical companies, and scientists from academia, the U.S. Food and Drug Administration (FDA), the European Medicines Agency (EMA), the National Institute of Neurological Disorders and Stroke (NINDS), and the National Institute on Aging (NIA). CPAD currently includes over 200 scientists, drug development and regulatory agency professionals, from member and non-member organizations. The data available in the CPAD database has been volunteered by CPAD member companies and non-member organizations.

\section{Acknowledgements}
\label{sec:acknowledgements}

Data collection and sharing for this project was funded in part by the Alzheimer's Disease Neuroimaging Initiative (ADNI) (National Institutes of Health Grant U01 AG024904) and DOD ADNI (Department of Defense award number W81XWH-12-2-0012). ADNI is funded by the National Institute on Aging, the National Institute of Biomedical Imaging and Bioengineering, and through generous contributions from the following: AbbVie, Alzheimer’s Association; Alzheimer’s Drug Discovery Foundation; Araclon Biotech; BioClinica, Inc.; Biogen; Bristol-Myers Squibb Company; CereSpir, Inc.; Cogstate; Eisai Inc.; Elan Pharmaceuticals, Inc.; Eli Lilly and Company; EuroImmun; F. Hoffmann-La Roche Ltd and its affiliated company Genentech, Inc.; Fujirebio; GE Healthcare; IXICO Ltd.; Janssen Alzheimer Immunotherapy Research \& Development, LLC.; Johnson \& Johnson Pharmaceutical Research \& Development LLC.; Lumosity; Lundbeck; Merck \& Co., Inc.; Meso Scale Diagnostics, LLC.; NeuroRx Research; Neurotrack Technologies; Novartis Pharmaceuticals Corporation; Pfizer Inc.; Piramal Imaging; Servier; Takeda Pharmaceutical Company; and Transition Therapeutics. The Canadian Institutes of Health Research is providing funds to support ADNI clinical sites in Canada. Private sector contributions are facilitated by the Foundation for the National Institutes of Health (\href{url}{www.fnih.org}). The grantee organization is the Northern California Institute for Research and Education, and the study is coordinated by the Alzheimer’s Therapeutic Research Institute at the University of Southern California. ADNI data are disseminated by the Laboratory for Neuro Imaging at the University of Southern California.

We would like to thank Pankaj Mehta and Alejandro Schuler for helpful comments while preparing the manuscript.

\bibliographystyle{plainnat}
\bibliography{ad}

\newpage
\appendix

\section{Expanded Methods}
\label{sec:methods}

\subsection{Data}%
\label{subsec:methods_data}

%%%
\begin{table*}[tp!]
\caption{Variables in the 3-month group $\subtmo {\v x}$, modeled by the 3-month component $\subtmo{p}$ only. See discussion in \subsecref{methods-problem-statement} for a description of the 3-month and 6-month components.}

{\renewcommand{\arraystretch}{0.7}%
\resizebox{\textwidth}{!}{
\begin{tabular}{|c|c|c|c|c|c|}
\hline
\textbf{Name}              & \textbf{Group} & \textbf{Category} & \textbf{Type} & \textbf{Longitudinal} & \textbf{Present {[}\%{]}} \\ \hline
ADAS cancellation          & 3         & Cognitive         & Ordinal       & Yes                   & 35.5                      \\ \hline
Serious adverse events     & 3         & Adverse events    & Ordinal       & Yes                   & 58.9                      \\ \hline
Heart rate                 & 3         & Clinical          & Continuous    & Yes                   & 98.5                      \\ \hline
Diastolic blood pressure   & 3         & Clinical          & Continuous    & Yes                   & 99.3                      \\ \hline
Systolic blood pressure    & 3         & Clinical          & Continuous    & Yes                   & 99.3                      \\ \hline
Weight                     & 3         & Clinical          & Continuous    & Yes                   & 83.3                      \\ \hline
Alanine aminotransferase   & 3         & Laboratory        & Continuous    & Yes                   & 52.7                      \\ \hline
Alkaline phosphatase       & 3         & Laboratory        & Continuous    & Yes                   & 52.7                      \\ \hline
Aspartate aminotransferase & 3         & Laboratory        & Continuous    & Yes                   & 52.7                      \\ \hline
Cholesterol                & 3         & Laboratory        & Continuous    & Yes                   & 41.8                      \\ \hline
Creatinine kinase          & 3         & Laboratory        & Continuous    & Yes                   & 47.5                      \\ \hline
Creatinine                 & 3         & Laboratory        & Continuous    & Yes                   & 52.7                      \\ \hline
Eosinophils                & 3         & Laboratory        & Continuous    & Yes                   & 42.4                      \\ \hline
Gamma-Glutamyl Transferase & 3         & Laboratory        & Continuous    & Yes                   & 39.6                      \\ \hline
Glucose                    & 3         & Laboratory        & Continuous    & Yes                   & 50.8                      \\ \hline
Hematocrit                 & 3         & Laboratory        & Continuous    & Yes                   & 45.9                      \\ \hline
Hemoglobin                 & 3         & Laboratory        & Continuous    & Yes                   & 52.6                      \\ \hline
Hemoglobin A1C             & 3         & Laboratory        & Continuous    & Yes                   & 36.7                      \\ \hline
Indirect Bilirubin         & 3         & Laboratory        & Continuous    & Yes                   & 36.4                      \\ \hline
Lymphocytes                & 3         & Laboratory        & Continuous    & Yes                   & 44.2                      \\ \hline
Monocytes                  & 3         & Laboratory        & Continuous    & Yes                   & 44.2                      \\ \hline
Platelet                   & 3         & Laboratory        & Continuous    & Yes                   & 52.4                      \\ \hline
Potassium                  & 3         & Laboratory        & Continuous    & Yes                   & 50.7                      \\ \hline
Sodium                     & 3         & Laboratory        & Continuous    & Yes                   & 46.1                      \\ \hline
Trisglycerides              & 3         & Laboratory        & Continuous    & Yes                   & 35.4                      \\ \hline
Region Europe              & 3         & Background        & Binary        & No                    & 95.7                      \\ \hline
Region Northern America    & 3         & Background        & Binary        & No                    & 95.7                      \\ \hline
Region Other               & 3         & Background        & Binary        & No                    & 95.7                      \\ \hline
\end{tabular}
}}
\label{tab:covariates-parent-only}
\end{table*}%
%%%

%%%
\begin{table*}[tp!]
\caption{Variables belonging to the 6-month group $\subsmo{\v x}$, modeled by the 6-month model $\subsmo{p}$. See discussion in \subsecref{methods-problem-statement} for a description of the 3-month and 6-month components.}

{\renewcommand{\arraystretch}{0.7}%
\resizebox{\textwidth}{!}{
\begin{tabular}{|c|c|c|c|c|c|}
\hline
\textbf{Name}              & \textbf{Group} & \textbf{Category} & \textbf{Type} & \textbf{Longitudinal} & \textbf{Present {[}\%{]}} \\ \hline
ADAS commands              & 3,6 & Cognitive         & Ordinal       & Yes                   & 99.7                      \\ \hline
ADAS comprehension         & 3,6 & Cognitive         & Ordinal       & Yes                   & 99.7                      \\ \hline
ADAS construction          & 3,6 & Cognitive         & Ordinal       & Yes                   & 99.6                      \\ \hline
ADAS delayed word recall   & 3,6 & Cognitive         & Ordinal       & Yes                   & 60.2                      \\ \hline
ADAS ideational            & 3,6 & Cognitive         & Ordinal       & Yes                   & 99.6                      \\ \hline
ADAS naming                & 3,6 & Cognitive         & Ordinal       & Yes                   & 99.6                      \\ \hline
ADAS orientation           & 3,6 & Cognitive         & Ordinal       & Yes                   & 99.6                      \\ \hline
ADAS remember instructions & 3,6 & Cognitive         & Ordinal       & Yes                   & 99.6                      \\ \hline
ADAS spoken language       & 3,6 & Cognitive         & Ordinal       & Yes                   & 99.7                      \\ \hline
ADAS word finding          & 3,6 & Cognitive         & Ordinal       & Yes                   & 99.7                      \\ \hline
ADAS word recall           & 3,6 & Cognitive         & Ordinal       & Yes                   & 99.6                      \\ \hline
ADAS word recognition      & 3,6 & Cognitive         & Ordinal       & Yes                   & 99.5                      \\ \hline
MMSE attention             & 3,6 & Cognitive         & Ordinal       & Yes                   & 60.8                      \\ \hline
MMSE language              & 3,6 & Cognitive         & Ordinal       & Yes                   & 57.9                      \\ \hline
MMSE orientation           & 3,6 & Cognitive         & Ordinal       & Yes                   & 60.8                      \\ \hline
MMSE recall                & 3,6 & Cognitive         & Ordinal       & Yes                   & 60.8                      \\ \hline
MMSE registration          & 3,6 & Cognitive         & Ordinal       & Yes                   & 60.8                      \\ \hline
CDR community              & 6          & Cognitive         & Ordinal       & Yes                   & 25.8                      \\ \hline
CDR home hobbies           & 6          & Cognitive         & Ordinal       & Yes                   & 25.8                      \\ \hline
CDR judgement              & 6          & Cognitive         & Ordinal       & Yes                   & 25.8                      \\ \hline
CDR memory                 & 6          & Cognitive         & Ordinal       & Yes                   & 25.8                      \\ \hline
CDR orientation            & 6          & Cognitive         & Ordinal       & Yes                   & 25.8                      \\ \hline
CDR personal care          & 6          & Cognitive         & Ordinal       & Yes                   & 25.8                      \\ \hline
Sex Female                 & 3,6 & Background        & Binary        & No                    & 100.0                     \\ \hline
Years of education         & 3,6 & Background        & Ordinal       & No                    & 23.0                      \\ \hline
Age                        & 3,6 & Background        & Continuous    & No                    & 99.4                      \\ \hline
Height                     & 3,6 & Background        & Continuous    & No                    & 80.0                      \\ \hline
AChEI/Memantine use    & 3,6 & Background        & Binary        & No                    & 100.0                     \\ \hline
History of hypertension    & 3,6 & Background        & Binary        & No                    & 99.8                      \\ \hline
History of type-2 diabetes & 3,6 & Background        & Binary        & No                    & 99.8                      \\ \hline
Clinical trial             & 3,6 & Background        & Binary        & No                    & 100.0                     \\ \hline
Amyloid status             & 3,6 & Background        & Binary        & No                    & 10.3                      \\ \hline
Vitamin B12                & 3,6 & Background        & Continuous    & No                    & 50.0                      \\ \hline
ApoE $\epsilon$4 allele count       & 3,6 & Background        & Ordinal       & No                    & 57.1                      \\ \hline
CSF phosphorylated tau 181 & 3,6 & Background        & Continuous    & No                    & 2.3                       \\ \hline
CSF total tau              & 3,6 & Background        & Continuous    & No                    & 3.5                       \\ \hline
\end{tabular}
}}
\label{tab:covariates-child}
\end{table*}%
%%%

Data for training and evaluating the model comes
from two different sources that provide complementary views of AD
disease progression. One source is the C-Path Online Data Repository
for Alzheimer\textquoteright s Disease (CODR-AD), a database provided
by the Critical Path for Alzheimer\textquoteright s Disease (CPAD) consortium \citep{romero2009coalition, neville2015development}, that consists of the control arms of 29 mostly mild to moderate
AD clinical trials with more than 7000 subjects. The other source
is from the Alzheimer\textquoteright s Disease Neuroimaging Initiative
(ADNI) database \citep{mueller2005ways}, a collection of four long-running observational studies
enrolling subjects across the AD disease spectrum since 2004, focusing
primarily on MCI and cognitively normal
subjects. These data have varying duration, visit interval, inclusion
criteria, and measured features. Notably, the ADNI studies typically
have a 6-month cadence and an extremely broad set of observations that include
imaging, biomarker data, and important disease severity measures; in contrast CODR-AD trials have a cadence of at most 3-months but a narrower set of measurements. The
CODR-AD data are encoded according to the Study Data Tabulation
Model (SDTM) format, a structured record format for clinical data
designed to facilitate review by regulatory authorities~\citep{kubick2007toward,hudson2018global}.  
The ADNI data are encoded in a custom format that accommodates the diverse array of measurements
made in the ADNI studies. These formats cannot be used for machine learning directly because they have complicated non-unique feature encoding schemes and contain extraneous information. We reprocessed the databases to extract measured features and encoded them into a consistent wide-form ("tidy") tabular format~\citep{wickham2014tidy}.

Sixty-four features were
selected for inclusion in the model as variables based on clinical
significance determined by recommendations from subject-matter experts and on the availability of data.   \Tabs{covariates-parent-only}{covariates-child} give basic properties of each of these variables.
Of primary interest for clinical trial applications are the components of ADAS-Cog, CDR, and MMSE, 
which are often used as inclusion criteria or clinical endpoints in trials (see Appendix~\subsecref{endpoints-covariates} for a summary of these endpoints). 
In addition, variables from the demographics, vital signs, medical history, questionnaires, laboratory
measurements, adverse events, and biomarker domains were also included. Each was classified
as a background (measured at baseline) or longitudinal (measured through time) variable and encoded as binary, ordinal,
categorical or continuous. Additionally, two binary features were
added, one indicating the baseline visit and another to denote whether the subject is enrolled in a clinical trial or an observational study. 

Studies were selected for inclusion in the processed dataset based on three criteria: the study must record measurements of ADAS-Cog (it is one of the most commonly used endpoints in AD clinical trials and it is crucial that we include it in our model), it must have a visit cadence of no more than 6 months (including studies with larger cadence would generate records with large missing portions, corresponding to the unrecorded visits) , and it must contain a disease population (we aim to model the progression of subjects with AD or MCI).
These criteria eliminate a number of CODR-AD studies that do not record ADAS-Cog data, the ADNI3 study that has a visit cadence of 12 months, and cognitively normal cohorts from the ADNI studies.  To differentiate cohorts with different disease severity in ADNI, the four ADNI studies were divided into smaller studies 
of cognitively normal, MCI, and AD subjects.

The resulting processed dataset consists of
6,919 subjects and 34,224 subject-visits split across
21 studies, with approximately 25\% of subjects in MCI stage and the remainder with AD. 

Before training, the dataset was split in the
ratio $0.5:0.2:0.3$ into training, validation, and test datasets, stratified by study.  
The test dataset was held out until the end of model development and used for all analyses shown in this work unless otherwise noted.

\subsection{Problem Statement}
\label{subsec:methods-problem-statement}

We denote the data as a set of vectors $\mathcal{D}=\{\v x^{(1)},\v x^{(2)},\cdots,\v x^{(N)}\}$,
where each measurement $\v x^{(i)}$ corresponds to a clinical record
of the $i$th subject, and the dataset contains $N$ subjects. Variables corresponding to the observed measurement are classified according to domain (continuous, binary, ordinal, categorical) which controls model encoding and whether they represent a static or time-dependent (longitudinal) observation \citep{walsh2020generating,fisher2019machine}. In the following, we denote the static portion of a trajectory as $\v x_s$ and the longitudinal portion $\v x_{1:T}$ with trajectory length $T$. Any combination of observations in $\v x$ may be missing, and are encoded by a special value.

We aim to train a model that represents the data distribution $p(\v x)$ to support conditional generation of synthetic clinical records. To simplify this task, we assume a causal Markov structure for $p(\v x)$ with a lag $L$,
\begin{equation}
p(\v x_s,\v x_{1:T})=p(\v x_s,\v x_{1:L})\prod_{t=L}^{T}p(\v x_{t}\mid\v x_{t-1:t-L},\v x_s), \label{eq:markovian-factorization}
\end{equation}
which naturally splits this model into a baseline component $p(\v x_\text{baseline} = (\v x_s, \v x_{1:L}))$ and an autoregressive component $p(\v x_{t}\mid\v x_{t-1:t-L},\v x_s)$. Following \citet{walsh2020generating}, we call a sample of a complete clinical record $\v x \sim p(\cdot)$ a Digital Subject; when conditioned on baseline measurements from an actual subject, $\v x_{L+1:T} \sim p(\cdot \mid \v x_\text{baseline})$, this clinical record is called a Digital Twin.

The training data contains some variables that are measured at a 3-month cadence and others that are measured at a 6-month cadence. We exploit this feature of the dataset by training one model for the 3-month variables $\subtmo{\v x}$ and another for the 6-month variables $\subsmo{\v x}$. Note that certain variables such as the ADAS-Cog endpoints are modeled by both components. Suppressing the time-dependence for the moment, the full model assumes a hierarchical form,
\begin{equation}
    p(\v x) \triangleq \subsmo{p}(\v x_\text{only 6mo} \mid \v x_\text{both 3 and 6 mo})\subtmo{p}(\subtmo{\v x}). \label{eq:conditional-objective}
\end{equation}
Here the 6-month model $\subsmo{p}$ explains variables that are observed only at 6-month intervals $\v x_\text{only 6mo}$ by conditioning on the overlap variables $\v x_\text{both 3 and 6 mo}$. The 3-month model $\subtmo{p}$ generates the remainder of the variables with a 3-month cadence. This simplifying assumption corresponds to neglecting the contribution from unobserved 3-month measurements to a subject's 6-month only variables.

We require that the trained model $p(\v x)$ supports the ability to generate Digital Twins. Using the hierarchical decomposition \eqref{conditional-objective}, this involves the following steps:
\begin{enumerate}
    \item Baseline measurements $\v x_\text{baseline}$ are taken directly from actual subjects that are to be simulated.
    \item The 3-month model is used to generate a partial trajectory of variables in 3-month group $\subtmo{\v x}$ by autoregressive sampling using \eqref{markovian-factorization}, conditioned on the baseline. 
    \item The 6-month model is used to complete the partial trajectory by autoregressive sampling of the 6-month variables, conditioned on the 3-month variables and the baseline variables.
\end{enumerate}
The component models must therefore support a variety of variable types, be tractable to train in the presence of missing data, and support conditional sequence generation. 

We choose CRBMs to represent both the 3-month and 6-month models because they meet these requirements and have been shown to be effective in modeling disease progression in AD~\citep{fisher2019machine} and MS~\citep{walsh2020generating}. Appendix~\secref{crbm-summary} contains a summary of CRBMs. For the 3-month model $\subtmo{p}$, a lag-2 CRBM with a cadence of 3 months is used. For the 6-month model $\subsmo{p}$, a lag-1 CRBM is used instead.

\subsection{Training Scheme and Hyperparameter Sweep}
\label{subsec:methods_training}

We train the composite model implicitly by maximizing the performance of the 3-month and 6-month CRBMs independently. This is equivalent to maximizing a composite pseudolikelihood~\citep{sutton2007piecewise} approximation to the joint model \eqref{conditional-objective}.

The 3-month component $\subtmo{p}(\subtmo{\v x})$ is a lag-2 model with 3 month visit cadence that is trained on 3-month variables $\subtmo{\v x}$ from the ADNI and CODR-AD datasets. Training samples for this model consist of the baseline variables and longitudinal variables of three consecutive visits with a spacing of 3 months. In the ADNI data (that have visit cadence of 6 months), these samples are of two types: either the first and last visit missing (type I), or the middle visit missing (type II). Instead of relying on the data imputation capabilities of the CRBM, we trained an auxiliary lag-1 CRBM imputation model with a visit cadence of 3 months on the CODR-AD data which is used to impute the type II samples in ADNI. Samples of type I consisting of mostly imputed data were not used during training to avoid introducing too much bias from the simpler imputation model.

The 6-month component $p_{6\text{month}}(\subsmo{\v x})$ is a lag-1 model with six month cadence that is trained on the 6-month variables $\subsmo{\v x}$ from ADNI and a single study from CODR-AD (that records CDR). As this is a simpler model, no special training procedures were required.

\begin{figure*}[tp!]
\includegraphics[width=6.5in]{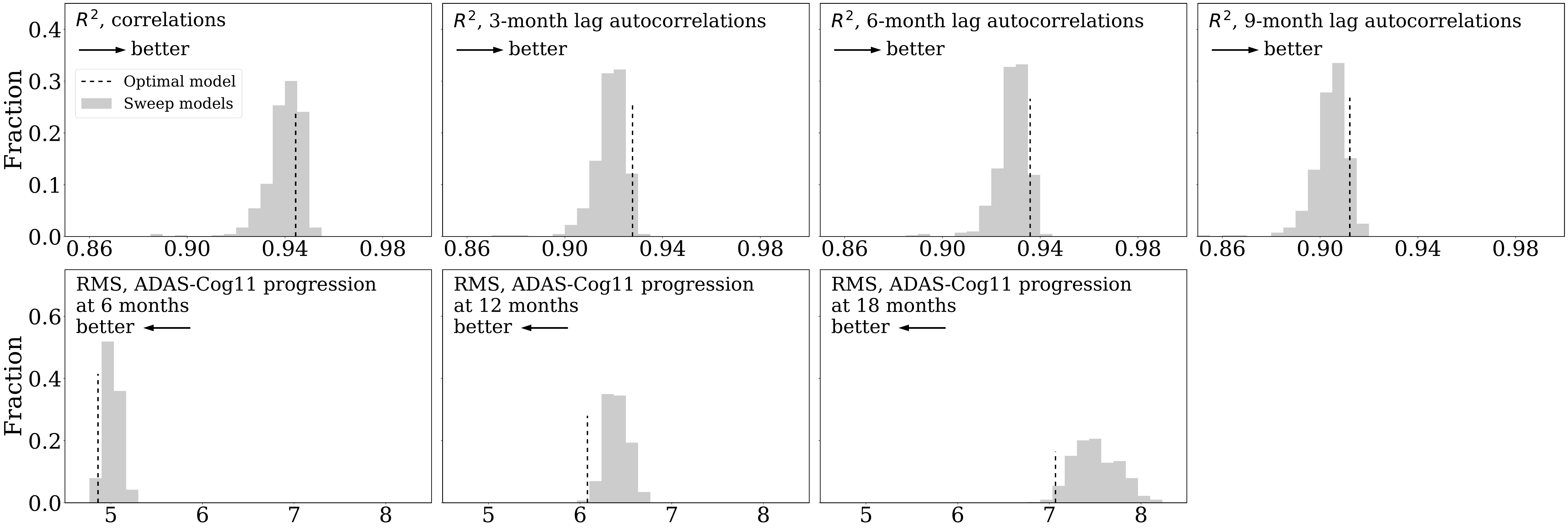}
\caption{{\bf Sweep distributions for the 3-month model.} The gray histograms represent the distributions of selection metrics (as reported in \Tab{selection_metrics}) over the hyperparameter sweep. The model selected from this sweep is shown as a dashed line at its value for each metric. For each metric, an arrow indicates whether smaller or larger values are better.                        
\label{fig:sweep_distributions_3months}}
\end{figure*}

\begin{figure*}[tp!]
\includegraphics[width=6.5in]{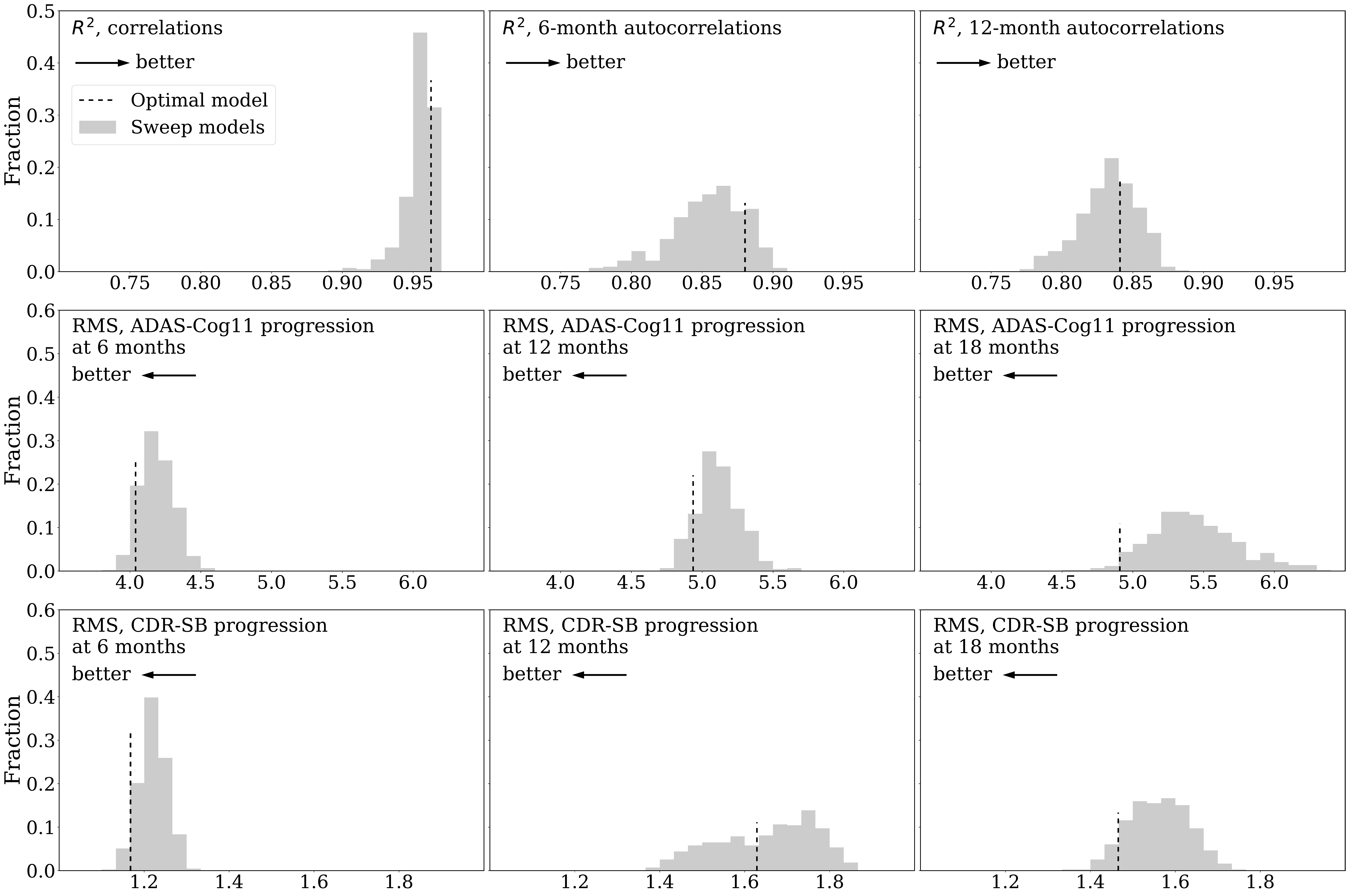}
\caption{{\bf Sweep distributions for the 6-month model.} Same as \Fig{sweep_distributions_3months} but for the 6-month model.
\label{fig:sweep_distributions_6months}}
\end{figure*}

The training procedure for each model is an elaboration of that described in Section IIC and Appendix C of \citet{walsh2020generating}, which we summarize here. We trained the 3-month and 6-month models by performing a grid search over hyperparameters in combination with minibatch stochastic gradient descent using the ADAM optimizer to obtain the parameters. We did not optimize the hyperparameters of the imputer model. We report details of the grids used for each model in \Tab{hyperparameter_grids}. For each grid point, we trained each model on the training portion of the dataset and evaluated it on the validation portion. We performed model selection with a two-step minimax procedure similar to the one outlined in \citet{walsh2020generating}. First, models from a grid search are ranked on the statistical and performance metrics of \Tab{selection_metrics} and assigned a score equal to the worst rank. Models are then re-ranked from minimum to maximum score and the bottom 75\% are rejected.  Second, we applied the same minimax procedure to the remaining 25\%, based on performance metrics only. Top-ranked models are selected as best and retrained on the joint training and validation splits. In \Fig{sweep_distributions_3months} and \Fig{sweep_distributions_6months} we show the distributions of selection metrics over all trained models and the values for the selected 3-month and 6-month models. We aimed to choose models that performed well across all metrics, even if they were not the best-performing on any single metric.

After an optimal 3-month and 6-month models were selected, each was retrained using these optimal hyperparameters on both the training and validation portions of the dataset.

\begin{table*}[tp!]
\caption{Hyperparameter grids used in model selection. The values selected by the minimax procedure are denoted in bold.}
\label{tab:hyperparameter_grids}
{\renewcommand{\arraystretch}{0.7}%
% \resizebox{\textwidth}{!}{
\begin{tabular}{|c|c|c|c|}
\hline
\multirow{2}{*}{\textbf{Hyperparameter}} & \multicolumn{3}{c|}{\textbf{Model}}                                                                                                                                                      \\ \cline{2-4} 
                                         & \textbf{3-month}                                                               & \textbf{6-month}                                                                        & \textbf{Imputer} \\ \hline
Batch size                               & \begin{tabular}[c]{@{}c@{}}\textbf{400}\\ 600\\ 800\end{tabular}                       & \begin{tabular}[c]{@{}c@{}}\textbf{100}\\ 200\\ 400\end{tabular}                               & \textbf{500}              \\ \hline
Number of epochs                         & \textbf{1000}                                                                          & \textbf{1000}                                                                                  & \textbf{1000}             \\ \hline
Learning rate                            & \begin{tabular}[c]{@{}c@{}}0.002\\ 0.004\\ 0.008\\ 0.012\\ \textbf{0.016}\end{tabular} & \begin{tabular}[c]{@{}c@{}}0.004\\ 0.008\\ 0.012\\ 0.016\\ 0.024\\ \textbf{0.032}\end{tabular} & \textbf{0.02}             \\ \hline
Beta std                                 & \begin{tabular}[c]{@{}c@{}}\textbf{0.15}\\ 0.30\\ 0.45\end{tabular}                    & \begin{tabular}[c]{@{}c@{}}\textbf{0.15}\\ 0.30\\ 0.45\end{tabular}                            & \textbf{0.15}             \\ \hline
Weight penalty                           & \textbf{0.001}                                                                         & \textbf{0.001}                                                                                 & \textbf{0.001}            \\ \hline
Monte Carlo steps                        & \textbf{25}                                                                            & \textbf{25}                                                                                    & \textbf{25}               \\ \hline
Adversary weight                         & \begin{tabular}[c]{@{}c@{}} 0\\ 0.15\\ \textbf{0.30}\end{tabular}                       & \begin{tabular}[c]{@{}c@{}}\textbf{0}\\ 0.15\\ 0.30\end{tabular}                               & \textbf{0.30}             \\ \hline
Number of hidden units                   & \begin{tabular}[c]{@{}c@{}}\textbf{65}\\ 130\\ 195\end{tabular}                        & \begin{tabular}[c]{@{}c@{}}\textbf{32}\\ 64\end{tabular}                                       & \textbf{32}               \\ \hline
\end{tabular}
}
\end{table*}

\begin{table*}[tp!]
\caption{Selection metrics for the 3-month and 6-month models. Statistical metrics measure how well the model reproduces correlations across variables, while performance metrics measure how well the model predicts the progression for key endpoints, i.e., ADAS-Cog11 and CDR-SB. $R^2$ is the coefficient of determination between data and predicted correlations or lagged autocorrelations. RMS is the root-mean-square error. We apply a two-step minimax selection process. First, models are ranked across statistical and performance metrics and assigned a score equal to the worst rank. Models are then re-ranked from minimum to maximum score and the top 25\% are selected.  Second, we applied the same minimax procedure to the pre-selected models, based on performance metrics only.
}
{\renewcommand{\arraystretch}{0.7}%
\begin{tabular}{|l|l|l|}
\hline
\textbf{Model} & \textbf{Statistical Metrics}                                                                                                                                     & \textbf{Performance Metrics}                                                                                                                                                                                                                                                        \\ \hline
3-month         & \begin{tabular}[c]{@{}l@{}}$R^2$ correlations\\ $R^2$ 3-month autocorrelations\\ $R^2$ 6-month autocorrelations\\ $R^2$ 9-month autocorrelations\end{tabular} & \begin{tabular}[c]{@{}l@{}}RMS ADAS-Cog11 progression at 6 months\\ RMS ADAS-Cog11 progression at 12 months\\ RMS ADAS-Cog11 progression at 18 months\end{tabular}                                                                                                                  \\ \hline
6-month          & \begin{tabular}[c]{@{}l@{}}$R^2$ correlations\\ $R^2$ 6-month autocorrelations\\ $R^2$ 12-month autocorrelations\end{tabular}                                  & \begin{tabular}[c]{@{}l@{}}RMS ADAS-Cog11 progression at 6 months\\ RMS ADAS-Cog11 progression at 12 months\\ RMS ADAS-Cog11 progression at 18 months\\ RMS CDR-SB progression at 6 months\\ RMS CDR-SB progression at 12 months\\ RMS CDR-SB progression at 18 months\end{tabular} \\ \hline
\end{tabular}
}
\label{tab:selection_metrics}
\end{table*}%

\subsection{Model Evaluation Methods}
\label{sec:model_evaluation_methods}
In this section we provide details on the model evaluation described in \subsecref{results_twins} and summarized in Figures~\ref{fig:zscores}, \ref{fig:moments}, and \ref{fig:adversary}.

In \Fig{zscores}, for each variable and each visit, we compare the values from a given subject to the distribution obtained from 100 Digital Twins of that subject. First, for a given variable $x_i$, for a subject $j$, and for visit $t$, we compute the p-value of the observation under the Digital Twins distribution 
\begin{equation}
    p_{i,j,t}(x_i^{\rm{data}}) = \Phi_{i,j,t}^{{\rm twins}}(x_i^{{\rm data}}),
\end{equation}
where $\Phi_{i,j,t}^{\rm{twins}}$ is the Digital Twins empirical cumulative distribution. If observed values are consistent with the model distributions one expects a uniform p-value distribution. Instead of testing directly this hypothesis, we define a derived statistic that is more interpretable. From a p-value we calculate a statistic from the inverse normal distribution,
\begin{equation}
    \varphi_{i,j,t} = \Phi^{-1}(p_{i,j,t}).
\end{equation}
The values are expected to be normally distributed across subjects with zero mean and unit standard deviation, if the observations are consistent with the model distributions. Means above (below) 0 indicate a biased model distribution with lower (higher) mean than the mean of the data. Variances smaller (larger) than 1 indicate a model variance larger (smaller) than variance of the data. In \Fig{zscores} we plot the mean and standard deviation of $\varphi$ for all variables and visits. We compute a 1-sample Kolmogorov-Smirnov test comparing $\mathcal{N}({\rm E}[\varphi_{i,j,t}], {\rm Var}[\varphi_{i,j,t}])$ and $\mathcal{N}(0, 1)$. The significance of the test is adjusted for multiple comparisons with a Bonferroni correction. The significance level is rescaled by the total number of comparisons, which is given by the number of variables multiplied by the number of visits. This leads to a corrected significance level of $0.05/126$.

In \Fig{moments} we compare summary statistics from test subjects and their Digital Twins. Let $\mathbf{x}_t^k$ for $k={\rm data}$ or $k={\rm twins}$ indicate either test subjects data or their Digital Twins at visit $t$, respectively. We measure means for each variable at each visit $\mu_t^k={\rm E}[\mathbf{x}_t^k]$, standard deviations for each variable at each visit $\sigma_t^k={\rm Var}[\mathbf{x}_t^k]^{1/2}$, and correlations over all times, $C^k_l={\rm Cov}[\mathbf{x}_t^k, \mathbf{x}_{t+l}^k]$, for $l=0,1,2,3$, which correspond to equal-time correlations and 3-, 6-, 9-months lagged auto-correlations, respectively. Each statistic is calculated over test subjects and their corresponding Digital Twins, where a single Digital Twin is assigned to each test subject. We compute a regression between the values obtained from test subjects and values obtained from Digital Twins, weighted by the fraction of data present for each particular point. Means and standard deviations span multiple orders of magnitude and ordinary least squares regression is particularly sensitive to outliers, so we use Theil-Sen regression. For correlations and auto-correlations we use ordinary least squares regression and we also report the corresponding coefficient of determination.

In \Fig{adversary} we train a classifier to distinguish actual subjects from Digital Twins, based on the complete set of longitudinal variables. We compare subjects with their Digital Twins at each visit, and we also consider a different classifier where the features are the differences of variables between two consecutive visits. Missing data is a potential source of bias, since actual subjects have missing values and their Digital Twins do not. We remove the potential bias from missing data by mean-imputing all missing values and by assigning the same imputed value to the corresponding Digital Twins. Classifiers are evaluated using 5-fold cross validation. For each fold, a model is trained on 4 folds and evaluated on the remaining one. Performance is evaluated as the average over the 5 folds. We generate 100 Digital Twins for each subject, and a separate classifier model is trained and evaluated for each set of twins. In the top panel of \Fig{adversary}, for any given visit, we report the average performance and the corresponding 95\% confidence interval error over the 100 trials. In the bottom panel, we report boxplots for the distribution of the logistic regression coefficients. The distribution includes all weights estimated from the 100 Digital Twins, from the 6 visits, and from the 5 folds of cross validation. Weights are related to the relative importance of features, where a weight equal to 0 corresponds to an irrelevant feature.

\section{ADAS-Cog, CDR, and MMSE variables}
\label{subsec:endpoints-covariates}

The ADAS-Cog test is widely used in clinical trials to asses cognitive decline in subjects with Alzheimer's Disease. ADAS-Cog most commonly includes between 11 and 14 tasks, which are scored independently and then combined into composite scores (e.g., ADAS-Cog11 or ADAS-Cog14). The tasks include both subject-completed tests and observer-based assessments, and evaluate the domains of memory, language, and praxis. 

ADAS-Cog is less suitable to detect changes at milder stages of dementia, and more sensitive tests have been introduced for this purpose. A widely used test is CDR. It employs an interview format to collect detailed information on the subject’s ability to function in various domains. It consists of 6 components which are commonly combined into the composite CDR Sum-of-Boxes (CDR-SB). CDR-SB is the gold standard for staging dementia in subjects who eventually develop AD and this is reflected in the early stage AD trials that have been run in the last decade.

A commonly used quick assessment of cognitive impairment is the MMSE questionnaire. MMSE has been a staple for inclusion and exclusion criteria in clinical trials for over 20 years and is used in nearly 100\% of trials for treatments from early to late-stage AD.

%%%

\section{Summary of CRBMs}
\label{sec:crbm-summary}

Restricted Boltzmann Machines \citep{ackley1985learning,smolensky1986,welling2004exponential} are a well-known type of generative latent-variable model that possess a number of qualities that make them suitable for modeling clinical data. An RBM is defined by an energy function \citep{lecun2006tutorial} $U(\v x,\v z)$ that measures the
compatibility between a set of observed variables $\v x$ and a set of
latent variables $\v z$ introduced to model the complex dependencies
in the data. The data distribution is then the marginal of a joint
distribution $p(\v x,\v z)$ defined by this energy function,
\begin{align*}
p(\v x) & \triangleq Z^{-1}\int \dif{\v z} \exp\left(-U(\v x,\v z)\right),
\end{align*}
with $Z$ the normalizing factor. This model makes few assumptions on
the domain of the variables $(\v x,\v z)$, which allows flexible
modeling of the data $\v x$ by choosing an appropriate energy function. By construction, RBMs can efficiently generate samples conditioned on observed data through the
use of MCMC methods, and can therefore be used for data imputation. They are also efficient to train, and naturally support training in the presence of missing observations. See recent reviews \citep{fischer2012introduction,hinton2010practical} for methods of training and sampling from RBMs.

A Conditional Restricted Boltzmann Machine (CRBM) is an extension of an RBM to support sequence data in the form of \eqref{markovian-factorization}. In the original development \citep{taylor2007modeling,mnih2012conditional}, the RBM was extended to model conditional distributions of the form $p(\v x_t \mid \v x_{t-1})$. Although sufficient for autoregressively extending sequences into the future, such a model cannot handle missing data in the conditioning set during training, and is hard to use for imputing backwards in time, $p(\v x_{t-1} \mid \v x_t)$. We use instead the CRBM architecture of \citet{fisher2019machine}, which differs from that in~\citep{taylor2007modeling,mnih2012conditional} because it is bidirectional and it allows for some of the visible variables (such as sex or race) to be time-independent ($\v x_s$). To make the model bidirectional, we define a single lag-$L$ joint distribution $p(\v x_{t:t+L},\v x_s)$ that models correlations between $L+1$ consecutive longitudinal measurements and the baseline variables $\v x_s$ as an RBM. This leads to a joint probability distribution given by
\begin{equation}
p(\v x_{t:t+L}, \v x_s) = Z^{-1} \int \dif \v z \, e^{-U(\v x_{t:t+L}, \v x_s, \v z)} \,,
\end{equation}
in which the energy function $U(\cdot)$ takes the form,
\begin{equation} \label{eq:probdist}
U(\v x_{t:t+L}, \v x_s, \v z)  = \sum_{i={0,1,\dotsc, L,s}}\biggl[\sum_{j} a_{i,j}(x_{t+i,j}) + \sum_{j \mu} 
W_{i,j \mu} \frac{x_{t+i,j}}{\sigma_{i,j}^2} \frac{z_{\mu}}{\epsilon_{\mu}^2} \biggr] + \sum_{\mu} b_{\mu} (z_{\mu}) \,.
\end{equation}
Each variable of the visible and latent layers has bias parameters determined by the choice of functions $a_{ij}(\cdot)$ and $b_\mu(\cdot)$, as well as scale parameters $\sigma_{ij}$ and $\epsilon_\mu$.  The connection between the layers is parameterized by the weight matrices $W_{i,j\nu}$. Understood as a conventional RBM, our CRBM contains the visible units for multiple time points coupled to a standard hidden layer. Clinical trajectories may be obtained from a CRBM model by sampling according to the Markov decomposition \eqref{markovian-factorization}. This is efficient because the joint RBM model $p(\v x_s, \v x_{t:t+L})$ is easy to sample conditionally, $\v x_{t+L} \sim p(\cdot \mid \v x_{t:t+L-1}, \v x_s)$.

To optimize the parameters of a CRBM, we replace the maximum likelihood objective (Eq \eqref{markovian-factorization}) with a piecewise pseudolikelihood approximation~\citep{sutton2007piecewise}, to obtain a piecewise loss $\mathcal L_\text{PL}$ in log space,
\begin{equation}
    \mathcal L_\text{PL} = \sum_{t=1,\dots,T-L} \log p(\v x_s, \v x_{t:t+L})
\end{equation}
that is averaged over adjacent windows of trajectories $(\v x_s, \v x_{t:t+L})$ which we call \emph{shingles}. This objective is then optimized with persistent contrastive divergence algorithm \citep{tieleman2008training}.

Training directly with this objective can result in poor sample quality. This is because maximum likelihood objectives allow the model to generate arbitrarily poor out-of-distribution data~\citep{theis2016note,fisher2018boltzmann}. We mitigate this problem by augmenting the loss with a weighted adversarial
(contrastive) loss term $\mathcal L_\text{adversary}$ that acts as a regularizer \citep{fisher2018boltzmann} (see \citet{chen2020simple} for a similar application in computer vision). We call the resulting training objective BEAM~\citep{fisher2018boltzmann}, 
\begin{equation}
    \mathcal L_\text{BEAM} = (1 - \lambda) \mathcal L_\text{PL} +        \lambda \mathcal L_\text{adversary},
\end{equation}
where $\lambda$ is a regularization strength hyperparameter. Although this procedure is superficially similar to generative adversarial network (GAN) training, we emphasize that the BEAM objective and model properties of CRBMs are fundamentally different from that of GANs. Indeed, we treat the weight of the regularization term as a hyperparameter which can be set to zero, whereas a GAN cannot
be trained without an adversary.

\begin{comment}
\label{subsec:Differences-of-BEAM}

\begin{itemize}
\item We learn the distribution intrinsically, GAN does not
\item We have stochastic mappings for the embedding space, GAN is deterministic
\item Adversary is an essential component of GAN training, but for us it
acts as a regularizer, a Contrastive Learning objective. In sweeps
we occasionally set the regularization parameter to zero.
\end{itemize}
\end{comment}

The computational cost of the BEAM procedure is similar to conventional RBM training and allows reusing existing RBM implementations. We refer the reader to Section IIC of \citet{walsh2020generating} for details on how CRBMs are trained on sequence data under the Markov assumption, and to \citet{fisher2018boltzmann} for a motivation for the BEAM objective.

\section{Additional Results on Clinical Endpoints Progression}
\label{sec:additional-results-endpoints}
In \Tab{fit_params} we report results from a linear regression of progression of test subject against the model predictions for various endpoints. We report slope, intercept, and Pearson correlation for ADAS-Cog11, MMSE, and CDR-SB for two representative visits at 12 months and 18 months from baseline. High correlations support the conclusion that our model predicts well progression also at subject level.
%%%

%%%
\begin{table*}[]
\caption{Fit parameters for subject-level score progression predictions. We fit a linear regression where subject progressions are regressed on the model predictions and report slope, intercept and their standard error in parenthesis. The model prediction for each subject is averaged over 100 Digital Twins generated for that subject. We also report the Pearson correlation coefficient between test subject progressions and the predicted progressions.}
{\renewcommand{\arraystretch}{0.7}%
\begin{tabular}{|c|c|c|c|}
\hline
\textbf{Score progression} & \textbf{Slope} & \textbf{Intercept} & \textbf{\begin{tabular}[c]{@{}c@{}}Pearson\\ correlation\end{tabular}} \\ \hline
\multicolumn{4}{|c|}{\textbf{12 months}}                                                                                                   \\ \hline
ADAS-Cog11                 & 0.70 (0.05)    & 0.68 (0.22)        & 0.36                                                                    \\ \hline
MMSE                       & 0.72 (0.06)    & -0.27 (0.15)       & 0.41                                                                    \\ \hline
CDR-SB           & 0.42 (0.08)    & 0.45 (0.10)        & 0.26                                                                    \\ \hline
\multicolumn{4}{|c|}{\textbf{18 months}}                                                                                                   \\ \hline
ADAS-Cog11                 & 0.91 (0.06)    & 0.46 (0.40)        & 0.49                                                                    \\ \hline
MMSE                       & 0.79 (0.08)    & -0.46 (0.27)       & 0.45                                                                    \\ \hline
CDR-SB           & 0.76 (0.13)    & 0.31 (0.23)        & 0.47                                                                    \\ \hline
\end{tabular}
}
\label{tab:fit_params}
\end{table*}
%%%

\section{Additional Results on Marginal Distributions}
\label{sec:additional-results-marginals}
In \Tabs{old_model_marginals}{old_model_marginals_2} we calculate means and standard deviations of marginal distributions for all longitudinal variables and compare test data with the model of this paper and the model of \citet{fisher2019machine}. We also report significant deviations of either models from data, using a t-test to compare means and the Levene’s test to compare standard deviations. We report 3 representative visits, but the test is performed for all visits from 3 months to 18 months in steps of 3 months, and significance is corrected for multiple comparisons for both models independently. We observe a larger number of significant deviations for MMSE marginals in the model of Fisher et al., showing that the model described in this paper improves predictions of MMSE. Both models show consistency for the remaining variables.

\begin{table*}[tp!]
\includegraphics[width=\columnwidth]{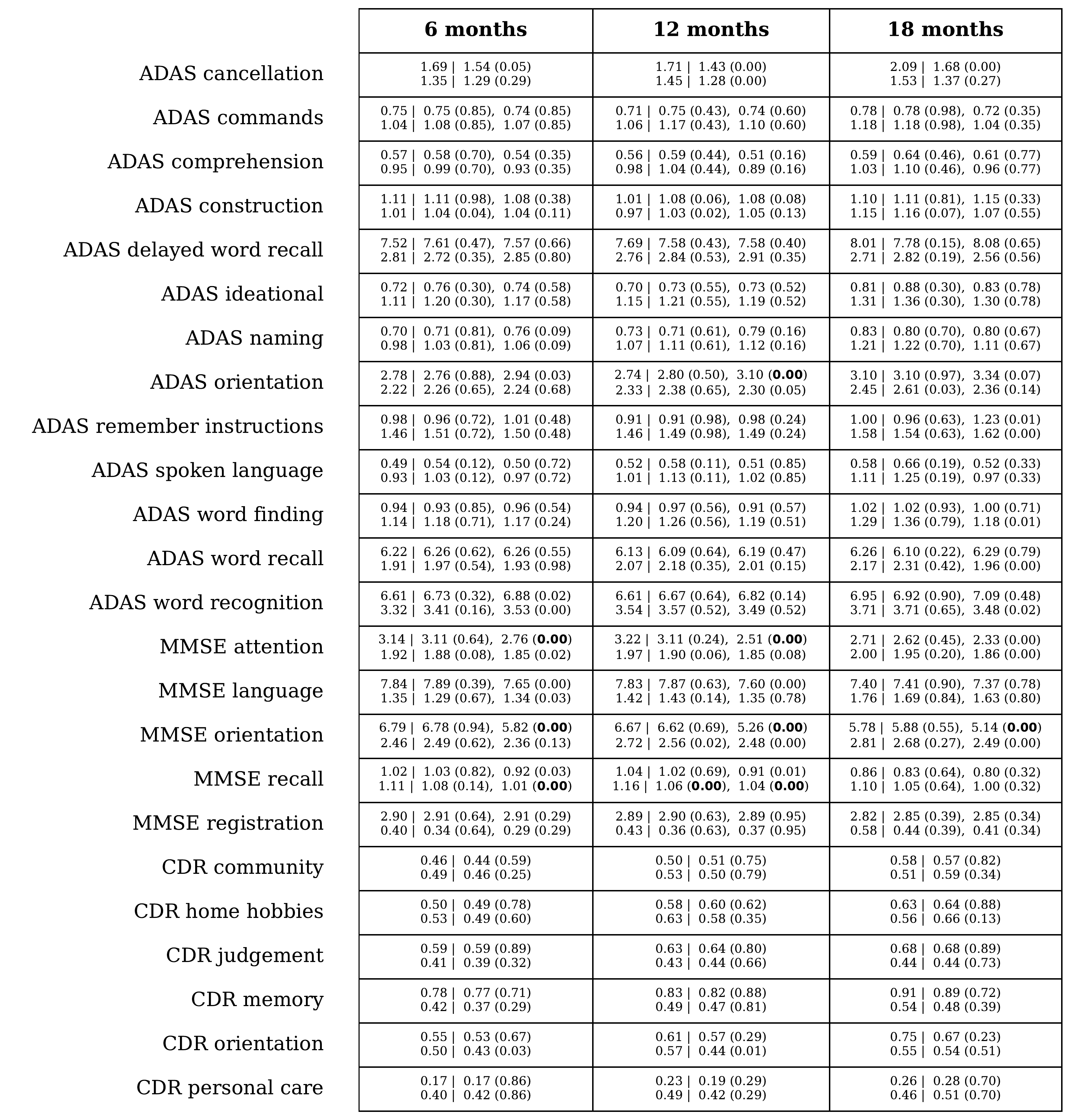}
\caption{Moments of marginal distributions (cognitive tests). For each variable and visit we show the mean (first row), and standard deviation (second row) of the marginal distributions. For each row, the first entry is from test subjects data, the second entry is from the model presented in this paper, and the third entry from the model of \citet{fisher2019machine}. We compare model means to data with a t-test, and standard deviations with Levene’s test, and report p-values in parenthesis (0.00 indicate values < 0.005). Bold values are significant after a Bonferroni correction, with significance level 0.05.
\label{tab:old_model_marginals}}
\end{table*}%

\begin{table*}[tp!]
\includegraphics[width=\columnwidth]{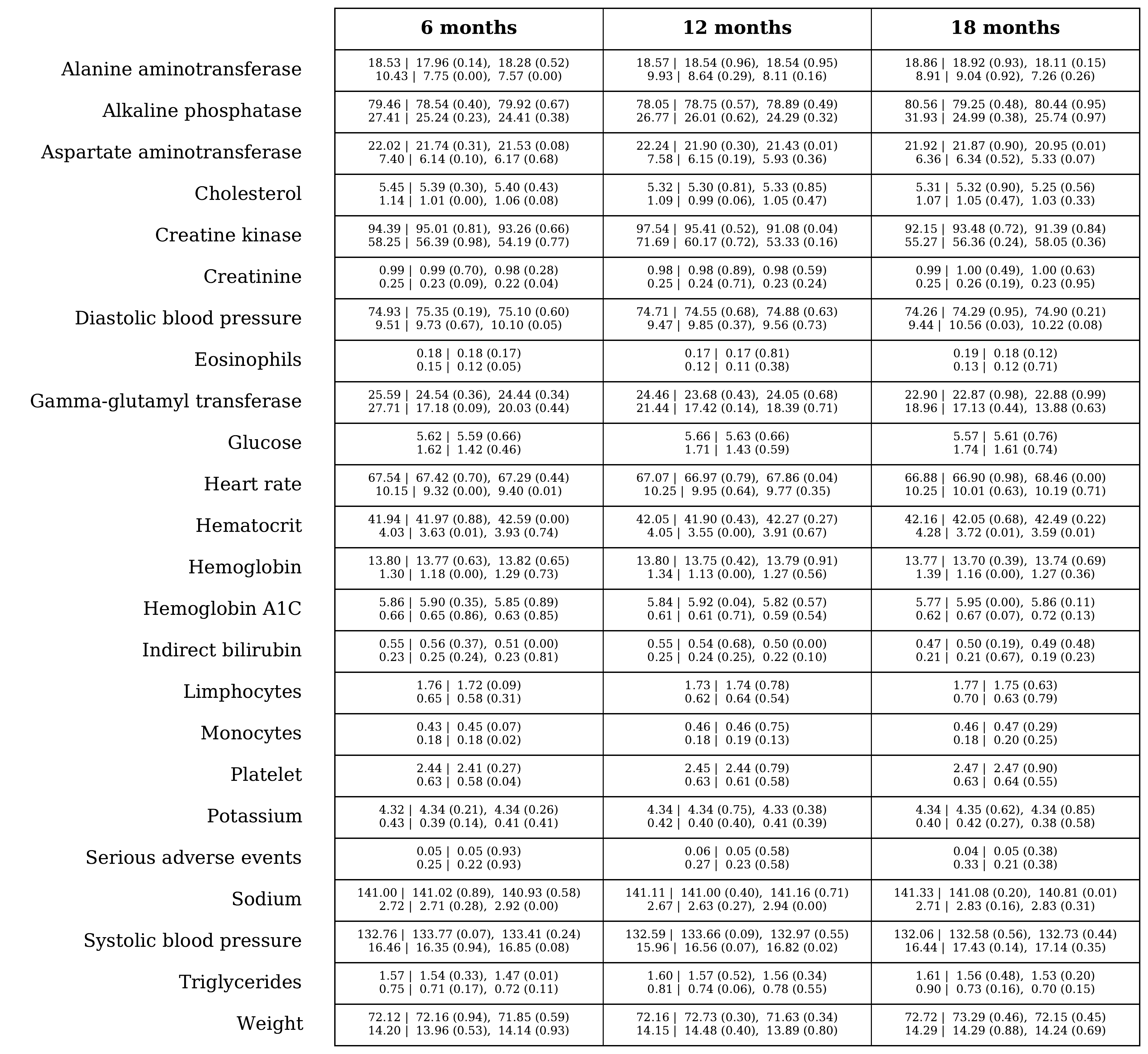}
\caption{Moments of marginal distributions (laboratory measurements). Same as \Tab{old_model_marginals} for laboratory measurements.
\label{tab:old_model_marginals_2}}
\end{table*}%

\end{document}